\documentclass[journal]{IEEEtran}
\ifCLASSINFOpdf
\else
\fi
\usepackage{graphicx}
\usepackage{amsmath}
\usepackage{algorithm}  
\usepackage{algorithmic} 
\usepackage{multirow}
\usepackage{amsfonts}
\usepackage{booktabs}
\usepackage[switch]{lineno}

\newcommand{\mG}{\mathcal{G}}

\begin{document}
%
\title{Bayesian Graph Convolutional  Network for Traffic Prediction}
%
%
%

\author{Jun Fu,
        Wei Zhou,~\IEEEmembership{Member,~IEEE}
        and~Zhibo Chen,~\IEEEmembership{Member,~IEEE}}

%
%

\markboth{Submission to IEEE Transactions on Intelligent Transportation Systems}%
{Shell \MakeLowercase{\textit{et al.}}: Bare Demo of IEEEtran.cls for IEEE Journals}
%



\maketitle

\begin{abstract}
	Recently, adaptive graph convolutional network based traffic prediction methods, learning a latent graph structure from traffic data via various attention-based mechanisms, have achieved impressive performance. However, they are still limited to find a better description of spatial relationships between traffic conditions due to: (1) ignoring the prior of the observed topology of the road network; (2) neglecting the presence of negative spatial relationships; and (3) lacking investigation on uncertainty of the graph structure. In this paper, we propose a Bayesian Graph Convolutional Network (BGCN) framework to alleviate these issues. Under this framework, the graph structure is viewed as a random realization from a parametric generative model, and its posterior is inferred using the observed topology of the road network and traffic data. Specifically, the parametric generative model is comprised of two parts: (1) a constant adjacency matrix which discovers potential spatial relationships from the observed physical connections between roads using a Bayesian approach; (2) a learnable adjacency matrix that learns a global shared spatial correlations from traffic data in an end-to-end fashion and can model negative spatial correlations. The posterior of the graph structure is then approximated by performing Monte Carlo dropout on the parametric graph structure. We verify the effectiveness of our method on five real-world datasets, and the experimental results demonstrate that BGCN attains superior performance compared with state-of-the-art methods.
\end{abstract}


%
\IEEEpeerreviewmaketitle

\section{Introduction}
\par Traffic congestion is a growing drain on the economy with the acceleration of urbanization. For example, the cost of traffic congestion in America reached \$124 billion in 2014, and will rise to \$186 billion in 2030, according to a report by Forbes~\cite{guerrini2014traffic}. One promising way to mitigate urban traffic congestion is to introduce Intelligent Transportation System (ITS). As an indispensable part in ITS, traffic prediction targets at predicting the future traffic conditions in road networks based on historical measurements. Accurate traffic prediction plays a vital role in traffic scheduling and management.
\begin{figure}
	\centering
	\begin{tabular}{c @{\hspace{.5em}} c}
		\includegraphics[width=.37\linewidth]{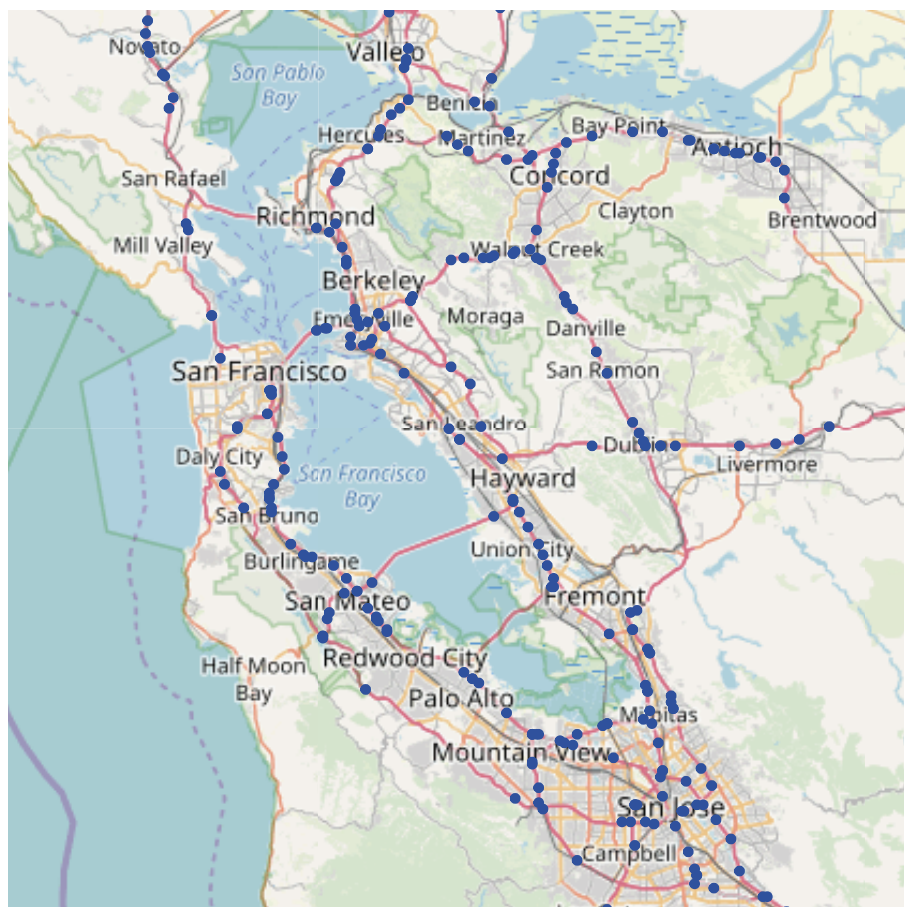} & 	\includegraphics[width=.45\linewidth]{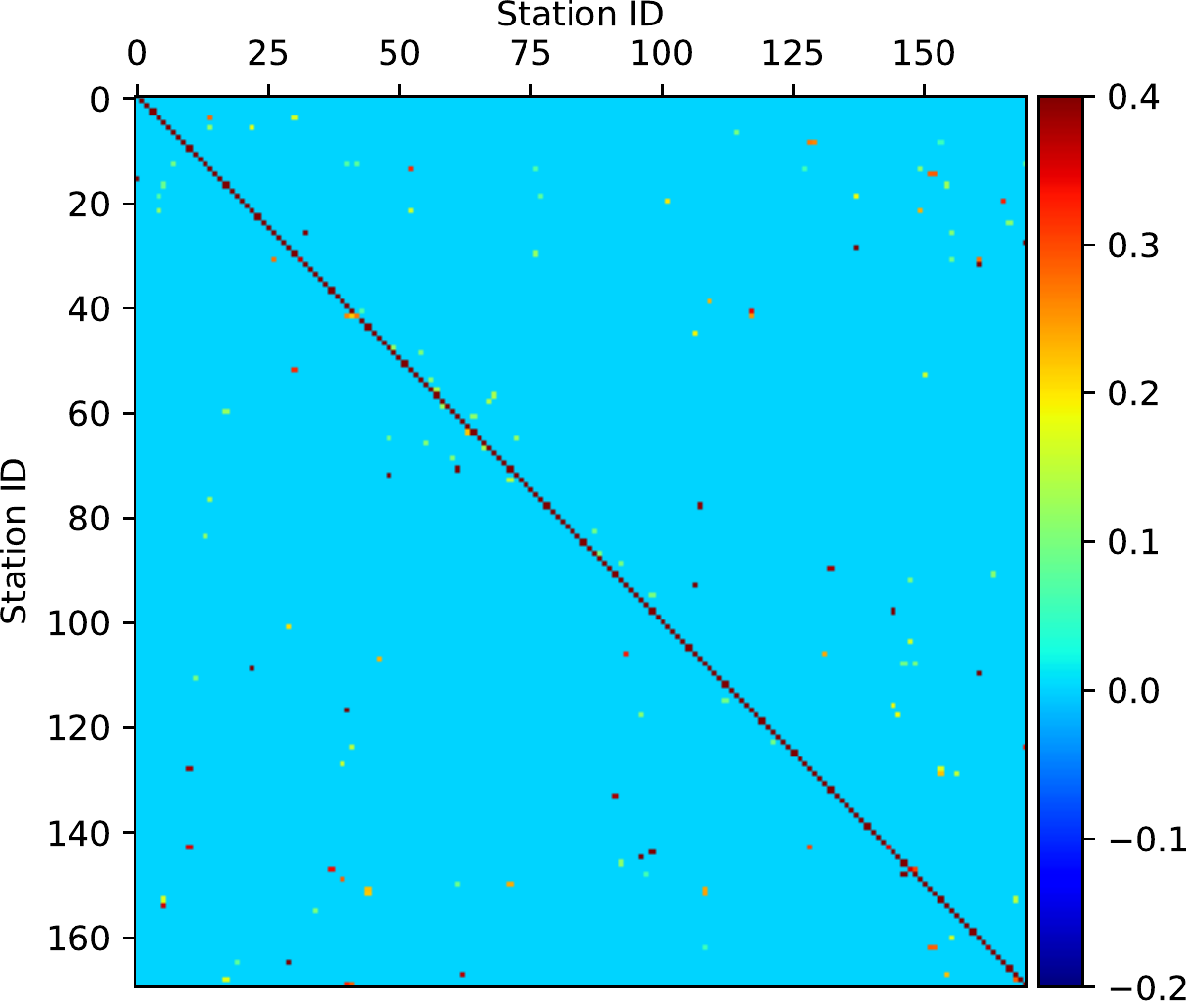}\\
		(a) & (b) \\
		\includegraphics[width=.45\linewidth]{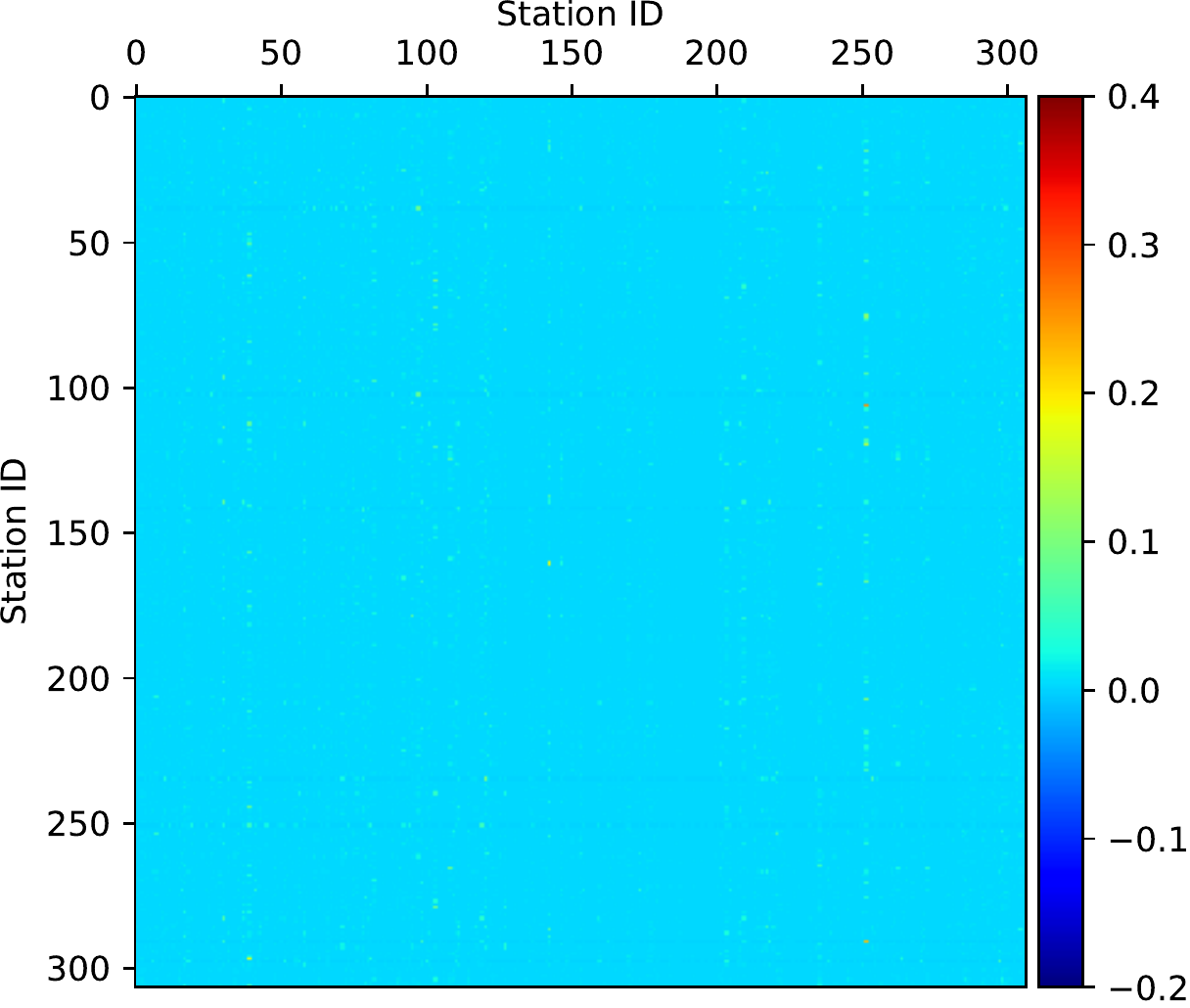} & 	\includegraphics[width=.45\linewidth]{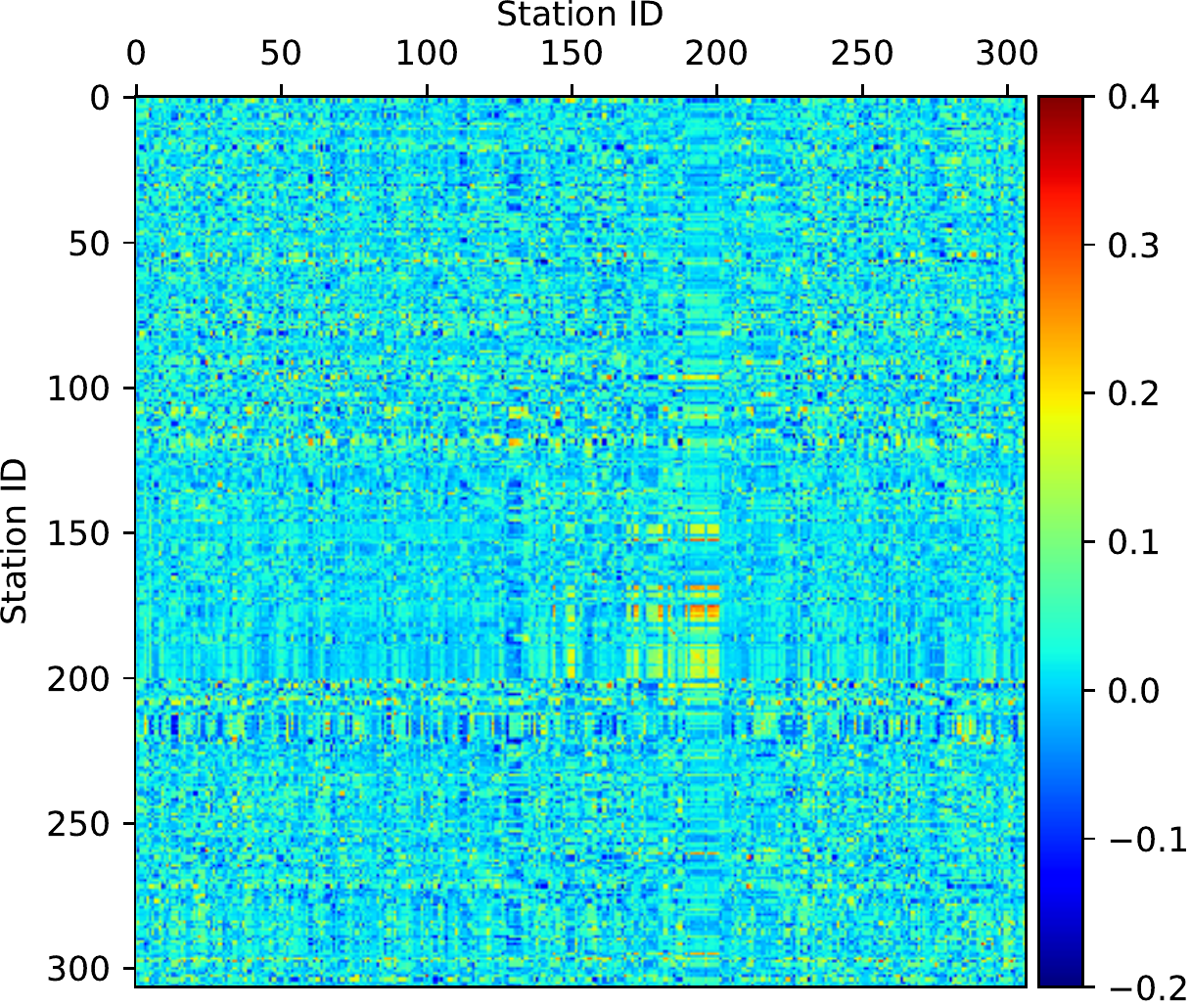} \\
		(c) & (d) \\
	\end{tabular}
	\caption{(a) PeMS sensor network in District 4 of California (b) Heuristically defined graph structure. (c) Learned graph structure by an attention-based method~\cite{wu2019graph}. (d) Learned graph structure by our proposed BGCN. Compared to the attention-based method, BGCN discovers more edges and includes both positive and negative spatial relationships. }
	\label{fig:first}
\end{figure}
\par Traffic forecasting is a challenging task due to the complex temporal dependencies (i.e, the traffic condition at a road is related to its historical observations) and spatial dependencies (i.e, the traffic conditions of adjacent roads influence each other). Traditional methods employ linear time series models~\cite{ahmed1979analysis,holden1995vector} for traffic forecasting. These methods fail to capture nonlinear temporal correlations, and ignore the presence of spatial dependencies. Recently, a broad of learning-based traffic predictors have been developed. They typically model temporal dependencies using recurrent neural networks (RNNs) or temporal convolution modules. As for spatial dependencies, they commonly deploy GCNs~\cite{kipf2016semi} because of the graph-structured road network. Despite the impressive results achieved, these graph-based methods are still limited to achieve more accurate prediction. This is mainly because the graph structure employed in GCNs is heuristically defined (i.e., roads are nodes, physical connections between roads are edges, and edge weights rely on the Euclidean distance between two roads), which may miss some important spatial correlations for traffic prediction (as shown in Fig.~\ref{fig:first}).
 
\par More recently, researchers turn to adaptive graph-based methods and focus on designing various attention mechanisms to learn the latent graph structure. However, existing adaptive graph-based methods suffer from the following three limitations. First, they learn the latent graph structure from scratch, ignoring the prior of the observed topology of the road network. Second, the attention-based graph learning methods tend to depress the negative spatial relationships due to the $SoftMax$ operation. However, some negative spatial relationships may be useful for traffic prediction. Third, introducing uncertainty into the graph structure (e.g., DropEdge~\cite{rong2019dropedge}) has proven its effectiveness in improve the generalization ability of GCNs. Unfortunately, GCN-based traffic predictors regard the graph structure as a deterministic variable, lacking investigation on uncertainty of the graph structure. 


\par The above concerns motivate us to propose a Bayesian Graph Convolutional Network (BGCN). BGCN considers the graph structure as a random sample drawn from a parametric generative model and infers the posterior of the graph structure based on the observed topology of the road network and traffic data. Specifically, BGCN first decomposes the graph structure into two parts: (1) a precomputed constant adjacency matrix that infers potential spatial relationships based on physical connections between roads using a Bayesian approach. (2) a learnable adjacency matrix that learns a  spatial pattern shared in all time steps from traffic data in an end-to-end manner and  can represent negative spatial correlations. Then, BGCN approximates the posterior of the graph structure by performing Monte Carlo dropout on the parametric graph structure.

The main contributions of our work lie in three folds:
\begin{itemize}
	\item This work is the first attempt to apply Bayesian graph convolutional networks in traffic prediction.
	\item This work presents an efficient approach to infer the posterior of the graph structure based on traffic data and the observed topology of the road network, which introduces little extra cost in computation and parameters.
	\item This work validates the effectiveness of the proposed BGCN on five real-world traffic datasets, and the experimental results show that BGCN surpasses state-of-the-art methods by a noticeable margin.
\end{itemize}
\par The rest of the paper is organized as follows. Section 2 reviews research works related to traffic prediction. Section 3 and 4 introduce details of our method and experimental results, respectively. Section 5 concludes this paper and points out some future directions.

\section{Related Work}
\subsection{Traffic prediction}
Traffic prediction is a classic space-time series forecasting problem. Traditional methods simply use time series models, e.g., ARIMA~\cite{ahmed1979analysis}, VAR~\cite{holden1995vector}, Support Vector Regression (SVR)~\cite{wu2004travel}, for traffic prediction. These methods fail to handle nonlinear temporal dynamics due to the stationary assumption. Moreover, they typically process each road individually, ignoring the spatial correlations between roads.

With the unprecedented success of deep learning, a variety of learning-based traffic predictors are proposed. Lv et al.~\cite{lv2014traffic} employ a stacked autoencoder to learn latent traffic flow features. Jia et al.~\cite{jia2016traffic} propose a Deep Belief Network for traffic speed prediction. Zhang et al.~\cite{zhang2016dnn} design DeepST, a convolutional neural networks (CNNs) based framework, to simultaneously model spatial-temporal correlations. Some RNN-based networks~\cite{yu2017deep,laptev2017time} are also proposed for traffic forecasting.  Despite achieving impressive performance in modeling temporal dynamics, they still have limited capabilities to model the spatial dependencies. This mainly because CNNs prefer to deal with grid-like data while the road network is naturally structured as a graph.

Recently, graph-based methods are proposed to address above concerns. Zhao et al.~\cite{zhao2019t} propose a temporal graph convolutional network (TGCN), which replace convolutional operations in RNNs with GCNs. Li et al.~\cite{li2017diffusion} regard the traffic flow as a diffusion process on a directed graph, and design a Diffusion Convolutional Recurrent Neural Network (DCRNN) for traffic forecasting. Yu et al.~\cite{yu2017spatio} introduce a Spatio-Temporal Graph Convolutional Network (STGCN), which leverages GCNs and gated CNNs for extracting spatial and temporal features respectively. Nevertheless, the graph structure in these GCN-based methods is manually defined and only reflects the topology of the road network. This may not be optimal for representing spatial correlations.

More recently, learning a latent graph structure for traffic prediction has attracted a lot of attention. Wu et al.~\cite{wu2019graph} explicitly learn representations of roads, and generate a static graph based on the similarity of two roads' representations. Guo et al. employ an attention mechanism~\cite{feng2017effective} to learn a data-dependent graph. Similar attention-based graph structure learning also is tried in STSGCN~\cite{song2020spatial}, and AGCRN~\cite{bai2020adaptive}. Unfortunately, in learning the underlying graph structure, these methods ignore the prior of the road network topology and lack investigation on the uncertainty of the graph structure.

\subsection{Graph Convolutional Network}
GCNs have been widely applied to a broad of applications, such as semi-supervised learning~\cite{jiang2019semi}, action recognition~\cite{yan2018spatial}, and quality assessment~\cite{xu2020blind,sun2021graphiqa}.  Recently, researchers shift attention to the uncertainty of the graph structure.  DropEdge~\cite{rong2019dropedge}, performing random dropout in the pre-defined graph structure, has proven  its effectiveness in enhancing generalization ability of GCNs. Bayesian Graph Convolutional Networks~\cite{zhang2019bayesian,pal2019bayesian}  not only consider the uncertainty of the graph structure and GCN weights, but also can derive the underlying connections between nodes. However, BGCNs are designed for semi-supervised node classification and have not been investigated in the more complicated spatial-temporal time series prediction.

\section{Method}
\subsection{Problem Definition}
 \noindent\textbf{Traffic Network}\quad  We assume the observed road network as a weighted directed graph $\mathcal{G}_{obs}=(\mathcal{V},\mathcal{E},\mathcal{A})$. Here, $\mathcal{V}$ is a set of $N = |\mathcal{V}|$ vertices, where vertex $v_i$  corresponds to the $i$-th road; $\mathcal{E}$ is a set of edges representing the connectivity between vertices; and $\mathcal{A} \in \mathbb{R}^{N \times N}$  is the weighted adjacency matrix, where $\mathcal{A}^{i,j}$  records the correlation between vertex $v_i$ and $v_j$.  The traffic condition at time step $t$ is regarded as a graph signal $X_t\in \mathbb{R}^{N \times D}$ on graph $\mathcal{G}_{obs}$, where $D$ represents the number of traffic characteristics (e.g., traffic flow, traffic speed, etc.). \\
\noindent\textbf{Problem}\quad  Given historical $T$ graph signals $\mathcal{X}=\{X_1,...,X_T\} \in \mathbb{R}^{N \times T \times D}$, we aim to  forecast $\tau$ future graph signals $\mathcal{Y}=\{{X}_{T+1},...,{X}_{T+\tau}\} \in \mathbb{R}^{N \times \tau \times D}$ using a mapping model $\mathcal{F}$:
\begin{equation}
\mathcal{Z} = \{\overline{X}_{T+1},...,\overline{X}_{T+\tau}\} = \mathcal{F}_{\theta}(X_1,...,X_T;\mathcal{G}_{obs}),
\end{equation}
where $\theta$ represents all the learnable parameters in the model $\mathcal{F}$.
\subsection{Motivation}
As aforementioned, the spatial correlation between traffic conditions is a key factor in traffic forecasting. Considering that the road network is naturally structured as a graph, existing works prefer to extract spatial features using a computation-friendly spectral graph convolution~\cite{kipf2016semi}:
\begin{equation}
f(X,\mathcal{A}) = \tilde{\mathcal{A}}XW,
\end{equation}
where $\tilde{\mathcal{A}} \in \mathbb{R}^{N\times N}$ is the normalized adjacency matrix with self-loops, $X \in \mathbb{R}^{N\times D} $ denotes the traffic condition at a certain time step, and $W \in \mathbb{R}^{D\times M} $ is the parameter matrix. To capture more meaningful graph representation, recent studies propose adaptive graph convolution, which learns a self-adaptive adjacency matrix in an attention-based mechanism:
\begin{equation}
\begin{split}
f(X,\tilde{\mathcal{A}}_{adp}) &= \tilde{\mathcal{A}}_{adp}XW, \\
\tilde{\mathcal{A}}_{adp} &= SoftMax(ReLU(E_1E^{T}_2)),
\end{split}
\end{equation}
where $E_1, E_2 \in \mathbb{R}^{N\times c}$ are two learnable node embeddings, and $\tilde{\mathcal{A}}_{adp} \in \mathbb{R}^{N\times N}_{+}$ is the generated adjacency matrix. 

However, such an adaptive graph convolution suffer from three limitations in the graph structure learning. First, it learns adjacency matrix from scratch, neglecting the prior of the observed topology of the road network. This may not be a good choice as the physical connection between roads is a vital clue for learning potential spatial correlations. Second, it is limited in modeling negative spatial correlations between traffic conditions because of the $SoftMax$ operation. Third, $\tilde{\mathcal{A}}_{adp} $ is still a deterministic variable, lacking investigation on the uncertainty of the graph structure.

\subsection{Bayesian Graph Convolutional Network}

We address above concerns with a Bayesian approach. Specifically, we regard the graph structure $\mathcal{A}$ as a sample drawn from a parametric generative model, and aim to infer the posterior predictive distribution :
\begin{equation}
p(\mathcal{Z}|\mathcal{X},\mathcal{Y},\mathcal{G}_{obs},W) = \int p(\mathcal{Z}|\mathcal{X}, W,\mathcal{G}) p(\mathcal{G}|\mathcal{X},\mathcal{Y},\mathcal{G}_{obs})d\mathcal{G},
\label{eq: posterior predictive distribution}
\end{equation}
where the item $p(\mathcal{Z}|\mathcal{X}, W, \mathcal{G})$ is modeled by a graph-based network, and the item $p(\mathcal{G}|\mathcal{X},\mathcal{Y},\mathcal{G}_{obs})$ aims to infer the posterior  of the graph structure using the train data $\{\mathcal{X},\mathcal{Y}\}$ and the observed topology of the road network $\mathcal{G}_{obs}$. It is worth noting that we only perform the posterior inference of the graph structure unlike conventional BGCNs. This is mainly because we find introducing uncertainty into the weights of GCNs has little impact on the traffic prediction in our experiment. 

In this paper, we perform the posterior inference of the graph structure in a coarse-to-fine fashion:
\begin{equation}
\begin{split}
p(\mathcal{Z}|\mathcal{X},\mathcal{Y},\mathcal{G}_{obs},\theta) &=\int p(\mathcal{Z}|\mathcal{X},W,\mathcal{G})p(\mathcal{G}|g,\mathcal{X},\mathcal{Y})\\
& \qquad\quad p(g|\mathcal{G}_{obs})d\mathcal{G}dg,
\end{split}
\label{eq:split}
\end{equation}
where we aim to first learn a coarse graph structure $g$ from the topology of the road network $\mathcal{G}_{obs}$,  and then learn a fine-level graph structure $\mathcal{G}$ based on $g$ and the paired traffic data $\{\mathcal{X},\mathcal{Y}\}$. Since there is no closed-form solution for the integral in Eq.~\ref{eq:split}, we introduce a Monte Carlo approximation:
\begin{align}
p(\mathcal{Z}|\mathcal{Y},\mathcal{X},\mG_{obs}) \approx 
\frac{1}{SC} \sum_{s=1}^S
\sum_{c=1}^{C} p(\mathcal{Z}|W,\mG_{s,c},\mathcal{X}),
\label{eq:MC_posterior}
\end{align}    
where $C$ weights $g_{c}$ sampled from $p(g|\mG_{obs})$, $S$ weights samples $\mG_{s,c}$ drawn from $p(\mathcal{G}|g_c,\mathcal{X},\mathcal{Y})$.  Consider sampling graph from $p(g|\mG_{obs})$ is time-consuming~\cite{pal2019bayesian},  we replace the integral over $g$ with a MAP process: 
\begin{equation}
\overline{g} = \mathop{\arg\max}_{g} \ \ p(g|\mathcal{G}_{obs}).
\label{bgcn:map}
\end{equation}
As described in the work~\cite{pal2019bayesian}, solving Eq.~\ref{bgcn:map} is equivalent to learning a $N \times N$ symmetric adjacency matrix of $g$:
\begin{equation}
\begin{split}
\mathcal{A}_{\overline{g}}= \mathop{\arg\min}_{\substack{\mathcal{A}_{g} \in \mathbf{R}_{+}^{N\times N}, \\ \mathcal{A}_{g}=\mathcal{A}_{g}^T }} \|\mathcal{A}_{g} \odot Z\|_1 -\alpha\mathbf{1}^T\log(\mathcal{A}_{g}\mathbf{1})  + \beta\|\mathcal{A}_{g}\|^2, 
\label{opt:adj_inference}
\end{split}
\end{equation}
where $\alpha$ and $\beta$ control the scale and sparsity of $A_{\overline{g}}$. Here, $Z\in\mathbb{R}^{N\times N}$ denotes the pairwise distance of roads in the embedding space:
\begin{equation}
Z_{p,q} = \|e_p - e_q \|^2,
\end{equation}
where $e_p$ and $e_q$ are the embedding vector of the $p$-th and $q$-th road. These embedding vectors of nodes are learned based on the physical connections between roads using  Graph Variational Auto-Encoder algorithm~\cite{kipf2016variational}.
After obtaining $Z$, we solve the Eq.~\ref{opt:adj_inference} via the prevalent optimization-based method~\cite{kalofolias2017large}. Then, we model the item $p(\mathcal{G}|g,\mathcal{X},\mathcal{Y})$ using a Monte Carlo approximation:
\begin{equation}
p(\mathcal{G}|g,\mathcal{X},\mathcal{Y}) = Dropout(\tilde{\mathcal{A}}_{\overline{g}}+\phi),
\label{eq:mc graph}
\end{equation}
where $\tilde{\mathcal{A}}_{\overline{g}}$ is the normalized adjacency matrix with self-loops, and $\phi \in \mathbb{R}^{N\times N}$ is a learnable adjacency matrix. We can notice that the graph structure comprises two parts: (1) a constant adjacency matrix $ \tilde{\mathcal{A}}_{\overline{g}} $ which offers some prior knowledge on spatial correlations; (2) a learnable adjacency matrix $ \phi $ which can include both positive and negative spatial correlations. Moreover, the graph structure is random due to the operation of $ Dropout $. Finally,  Eq.~\ref{eq:MC_posterior} is simplified as follows:
\begin{equation}
p(\mathcal{Z}|\mathcal{X},\mathcal{Y},\mathcal{G}_{obs},W) \approx \frac{1}{S} \sum_{s=1}^{S} p(\mathcal{Z}|\mathcal{X},W,\mG_s),
\label{bgcn:mc}
\end{equation} 
where $S$ weights samples $\mG_s$ drawn from $A_{\overline{g}} + \phi$ via dropout. The graph convolution in the item $  p(\mathcal{Z}|\mathcal{X},W,\mG_s)  $ is performed as follows:
\begin{equation}
\begin{split}
f(X, \tilde{\mathcal{A}}_{\overline{g}}, \phi) &= Dropout(\tilde{\mathcal{A}}_{\overline{g}}+\phi)XW.\\
\end{split}
\end{equation}
\begin{figure}[htbp]
	\centering
	\includegraphics[height=7cm]{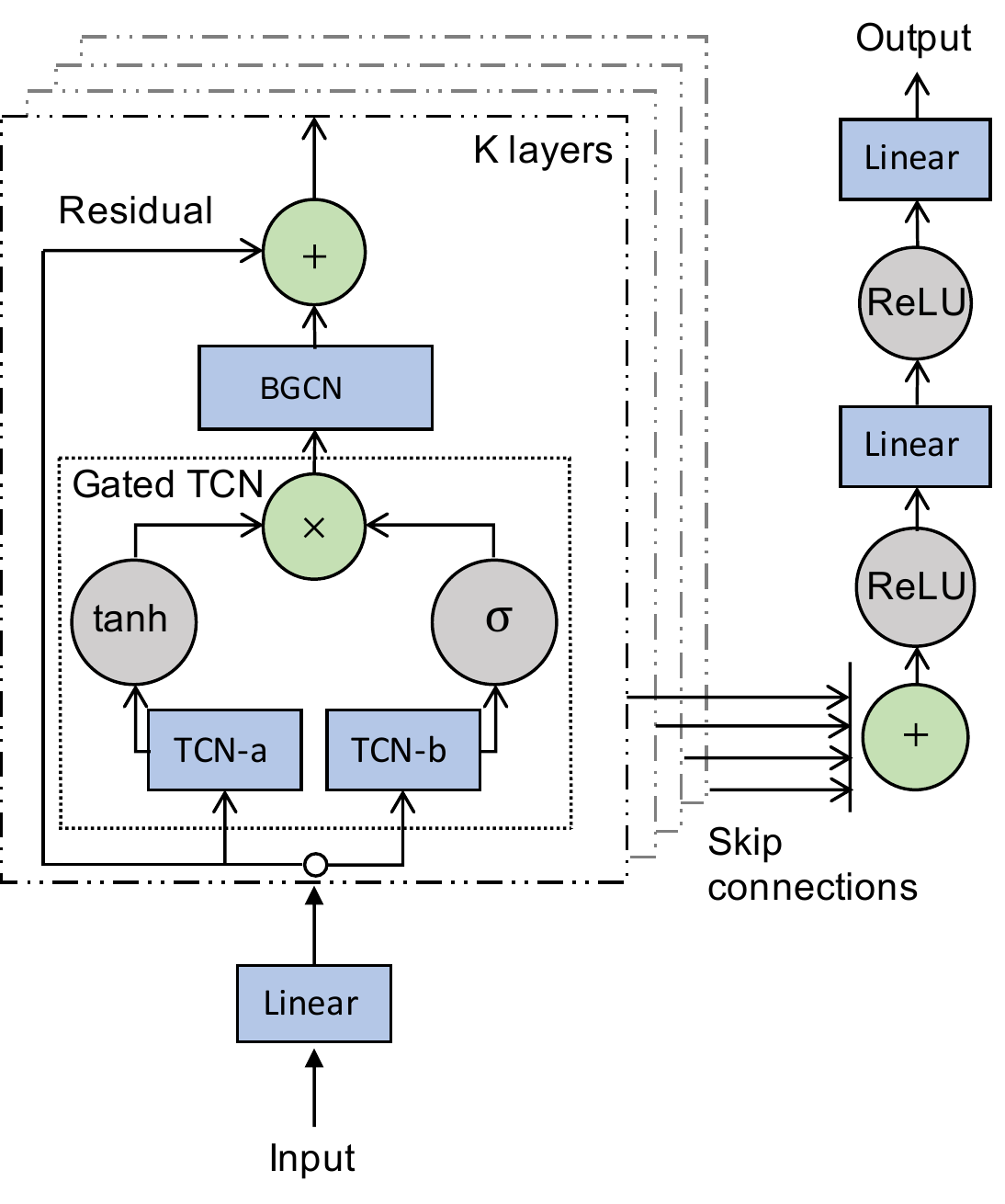}
	\caption{Framework of BGCN-applied Graph WaveNet. TCN-a and TCN-b are two types of temporal convolutional networks. Linear represents fully-connected layers. $\sigma$ denotes the sigmoid activation function.}
	\label{fig:BGCRN}
\end{figure}
\subsection{Network Architecture}
We adopt the architecture of Graph WaveNet. As shown in Fig.~\ref{fig:BGCRN}, It is composed of stacked spatial-temporal layers and an output layer. A spatial-temporal layer  consists of GCN and a gated temporal convolution layer (Gated TCN) that contains two parallel temporal convolution layers (TCN-a and TCN-b). We replace GCN in all spatial-temporal layers  with our proposed BGCN.

We choose to use mean absolute error (MAE) as the training objective, which is formulated as follows:
\begin{equation}
L = \frac{1}{\tau ND} \sum_{i=1}^{\tau}\sum_{j=1}^{N}\sum_{k=1}^{D} \lVert \overline{X}_{T+i}^{jk} - {X}_{T+i}^{jk}\rVert,
\label{eq:loss}
\end{equation}
where $\overline{X}_{T+i}$ and $ {X}_{T+i}$ are the predicted results and the ground truth at time step $T+i$. The whole training algorithm is described in Algorithm \ref{alg:bgcn_non_param}.

\subsection{Discussion}
In despite of considering the uncertainty, the BGCN~\cite{zhang2019bayesian} neglects the potential dependence of the graph structure on the training data. To address this issue, Pal et.al~\cite{pal2019bayesian} propose a non-parametric BGCN. The adjacency matrix discovered by the non-parametric BGCN relies on a neighborhood-based similarity rule. Moreover, it  is symmetric, which is easily violated in traffic prediction because the mutual influence of traffic conditions is not equal. Compared to the non-parametric BGCN, our proposed BGCN can learn the graph structure in an end-to-end manner. Furthermore, the learned graph structure can be either symmetric or asymmetric.
\begin{algorithm}[tb]
	\caption{Training methodology of BGCN}
	\label{alg:bgcn_non_param}
	\begin{algorithmic}[1] 
		\STATE \textbf{ Input:} $\mathcal{G}_{obs}$, training datasets $\mathcal{D}$
		\STATE \textbf{ Output:} $W$, $\mathcal{A}_{\overline{g}}$, $\phi$
		\STATE Randomly initialize $W$ 
		\STATE  Initial $\phi$: $\phi \leftarrow 1e^{-6}$
		\STATE  Obtain $\mathcal{A}_{\overline{g}}$ via solving Eq.~\ref{opt:adj_inference}
		\FOR{ $epoch=1$ {\bfseries to} $MaxEpoch$}
		\STATE Sample the graph structure $\mathcal{G}$ from $\mathcal{A}_{\overline{g}}+ \phi$ via dropout
		\STATE Sample a batch of data $\{\mathcal{X},\mathcal{Y}\}$ from $D$
		\STATE Obtain the predicted result $\mathcal{Z}$ 
		\STATE Optimize $W$ and $\phi$ by minimizing Eq.~\ref{eq:loss}.
		\ENDFOR
	\end{algorithmic}
\end{algorithm}

\begin{table}
	\caption{the details for the datasets.}
	\begin{tabular}{l|ccccc}
		\toprule 
		Dataset & PeMS3 & PeMS4 & PeMS7 & PeMS8 & PeMS-Bay \\
		\midrule
		Sensors & 358 & 307 & 883 & 170 & 325 \\
		Time steps & 26208 & 16992 & 28224 & 17856 & 52116 \\
		Mean & 181.40 &  207.25 & 309.59 & 229.99 & 62.75 \\
		STD & 144.41& 156.49 & 189.49 & 145.61& 9.37 \\
		Time interval & \multicolumn{5}{c}{5 minutes} \\
		Daily range & \multicolumn{5}{c}{00:00-24:00} \\
		\bottomrule
	\end{tabular}
	\label{tb:dt}
\end{table}
\section{Experiments}

\begin{table*}
	\centering
	\caption{performance comparison of different approaches on five datasets. ``*'' means reported results in the STSGCN.}
	
	\begin{tabular*}{\hsize}{@{}@{\extracolsep{\fill}}ll|ccccccccc@{}}
		
		\toprule
		
		\multirow{2}{*}{Dataset}	& \multirow{2}{*}{Metric}	& \multirow{2}{*}{HA} & \multirow{2}{*}{VAR}	& \multirow{2}{*}{GRU-ED} & \multirow{2}{*}{DCRNN*}& \multirow{2}{*}{STGCN} & \multirow{2}{*}{Graph WaveNet}& \multirow{2}{*}{STSGCN*}& \multirow{2}{*}{AGCRN}& \multirow{2}{*}{BGCN}\\
		
		& & & & & & & & & &\\
		
		\midrule
		
		\multirow{3}{*}{\rotatebox{0}{PeMS3}} & MAE   & 24.96	& 	-	  &		 19.90	&	18.03	 	& 	16.29	   &	14.66	 & 17.33	& 	15.70	 	&  \textbf{14.35}	\\
		
		& RMSE                     &  46.07 & 	-	  &		32.68 	&	30.06	& 	27.73	   &	 25.32	 	 & 28.65	& 		 27.71	&  \textbf{25.28}	\\
		& MAPE (\%)                      & 25.84 	 & 	-	  &		18.83  	&	18.09 	& 	17.80 	   &	15.29 	 	 & 16.58 	& 	14.82 	 	& \textbf{14.47}	\\
		
		\midrule
		
		\multirow{3}{*}{\rotatebox{0}{PeMS4}} & MAE   & 24.56	& 	22.82	  &	 25.63	 	&	24.48	 	& 	21.09	   &		 19.23 & 21.09	& 		 19.86	& \textbf{18.82}	\\
		
		& RMSE                     &  39.91 & 	35.26	  &		 39.84	&	37.86	 	& 	33.08	   &		 30.71	& 	33.45&	 32.00	& \textbf{30.34}	\\
		
		& MAPE (\%)                         & 16.56 	 & 	16.35 	  &		 16.41 	&	16.75 	 	& 	14.57 	   &	13.22 	 	& 	13.85 	 & {12.90} 	&  \textbf{12.87}	\\
		
		\midrule
		
		\multirow{3}{*}{\rotatebox{0}{PeMS7}} & MAE   & 28.49	& 	-	  &		27.20 	&	24.78	 	& 	22.63	   &	20.77	 	& 	24.12&	 21.81	&  \textbf{20.09}	\\
		
		& RMSE                     &  52.59 & 	-	  &		 43.13	&	37.88	 	& 	35.50	   &	33.39	 	& 	38.76	& 34.97	& \textbf{32.86}	\\
		
		& MAPE (\%)                        & 11.98 	 & 	-	  &		 11.56 	&	11.33 	 	& 	10.01 	   &	8.91 	 	& 	10.16 &	 9.18 	&  \textbf{8.45 }	\\
		
		\midrule
		
		\multirow{3}{*}{\rotatebox{0}{PeMS8}} & MAE   & 21.23	& 	19.87	  &		 20.10	&	17.83	 	& 	16.98	   &	15.43	 	& 	17.04&	 16.29	&  \textbf{14.65}	\\
		
		& RMSE                     &  36.72 & 	29.29	  &	 31.68	 	&  	 27.68	& 	26.58	   &		 	24.19 & 	26.62&	 25.66	&  \textbf{23.43}	\\
		
		& MAPE (\%)                        & 13.75 	 & 	13.04 	  &		12.33  	&	11.42 	 	& 	11.58 	   &	10.25 	 	& 	10.89 &	 10.32 	& 	\textbf{9.42} \\
		
		\midrule
		
		\multirow{3}{*}{\rotatebox{0}{PeMS-Bay}} & MAE   & 2.88	& 	2.24	  &		 1.96	&		- 	& 	1.77	   &		 1.65	& 	- &	 1.71	& \textbf{1.61}	\\
		
		& RMSE                     &  5.59 & 	3.97	  &		 4.69	&	-	 	& 	3.87	   &	3.66	 	&  -&		 3.88	& 	\textbf{3.63}\\
		
		& MAPE (\%)                        & 6.77 	 & 	 4.83 		  &		 4.47 	&		 -	& 	4.05 	   &		\textbf{3.65 }	& 	-&	  3.88 	& 3.71	\\
		
		\bottomrule
		
	\end{tabular*}
	\label{tb:all results}
	
\end{table*}

\subsection{Datasets}
\par To verify the effectiveness of the proposed method, we conduct experiments on five real-world traffic datasets: PeMS3, PeMS4, PEMS7, PeMS8, and PeMS-Bay. These  datasets are provided by Caltrans Performance Measure System (PeMS), which records the highway traffic in California every 30 seconds. The more details for the datasets are presented in Table~\ref{tb:dt}.\\
\noindent\textbf{PeMS3}: It refers to the traffic data collected by 358 loop detectors in District 3 of California from September 1st to November 30st in 2018.\\
\noindent\textbf{PeMS4}: It refers to the traffic data collected by 307 loop detectors in the San Francisco Bay Area from January 1st to February 28st in 2018. \\
\noindent\textbf{PeMS7}: It refers to the traffic data collected by 883 loop detectors in District 7 of California from May 1st to August 31st in 2017. \\
\noindent\textbf{PeMS8}: It refers to the traffic data gathered by 107 loop detectors in the San Bernardino Area from July 1st to August 31st in 2016. \\
\noindent\textbf{PeMS-Bay}: It refers to the traffic data gathered by 325 loop detectors in the Bay Area from January 1st to May 31st in 2017. \\

\subsection{Data preprocessing}
We adopt the following strategies to preprocess the traffic data, which is consistent with previous studies~\cite{song2020spatial,bai2020adaptive}. We aggregate the traffic data in a 5-minute interval. As a result, every loop detector contains 288 traffic data points per day.  We split all the traffic datasets into training sets, validation sets, and test sets in a ratio of 6:2:2. We discard the missing values and use Z-score method to normalize traffic data. 

Regarding  the observed adjacency matrix $\mathcal{A}_{obs}$, we construct it using the same method as DCRNN~\cite{li2017diffusion}:
\begin{equation}
	\mathcal{A}^{i,j}_{obs} = \begin{cases}
	\exp(-\frac{d^2_{i,j}}{\xi^2}), i\neq j \ \text{and} \  \exp(-\frac{d^2_{i,j}}{\xi^2}) \geq \epsilon,\\
	\\
	0, \text{otherwise}\\
	\end{cases}
\end{equation}
where $\mathcal{A}^{i,j}_{obs}$ is the edge weight between the $i$-th road and the $j$-th road, $d_{i,j}$ denotes the distance from the $i$-th road to the $j$-th road, $\xi$ is the standard deviation of distances, $\epsilon$ is a threshold and set as 0.1.

\subsection{Experimental Settings} 
\noindent\textbf{Implement Details} \quad We implement  BGCN based on the popular deep learning framework PyTorch ~\cite{paSZke2017automatic}, and conduct all experiments on a Linux server with one NVIDIA 1080Ti GPU card. The parameter setting of our framework keeps the same with Graph WaveNet. All the traffic datasets share the following training settings. Our goal is  to predict traffic conditions in the next hour based on the measurements of the past one hour. We set the Monte Carlo dropout probability to 0.5. We optimize all trainable variables with the Adam~\cite{kingma2014adam} optimizer for 100 epochs. The learning rate is initialized as 0.001 and decreases to 0.0001 at the 50-th epoch. During training, 64 pairs are randomly generated from the training dataset per iteration. \\\\
\noindent\textbf{Baselines} To evaluate the overall performance of our work, we compare BGCRN with the following baselines: 
\begin{itemize}
	\item Historical Average (HA). It considers the traffic conditions of each road as a seasonal process and uses the average of previous seasons as the prediction. The duration of a season is set to a week.
	\item Vector Auto-Regression (VAR)~\cite{holden1995vector}. It is a linear time series model, which can capture spatial correlations between traffic conditions.  We implement it based on \textit{statsmodel} python package. The number of lags for PeMS4, PeMS8, and PeMS-Bay is set to 8, 3, and 3, respectively. We do not report the performance on the PeMS3 and PeMS7 dataset due to the poor prediction results of VAR.
	\item GRU-ED. It adopts an encoder-decoder framework, where both encoder and decoder consist of a stacked GRU. The hidden size of GRU is set to 128.
	\item DCRNN~\cite{li2017diffusion}. Similar to GRU-ED, it also uses an encoder-decoder architecture, but equips both the encoder and decoder with diffusion convolution recursive layers for  capturing spatial-temporal features.
	\item STGCN~\cite{yu2017spatio}. It extracts spatial-temporal correlations using spatial-temporal graph convolutional networks, which is a combination of temporal gated CNNs and spatial GCNs. Different from the original STGCN, we implement the output layer to generate prediction for all horizons at one time instead of one horizon per time.
	\item Graph WaveNet~\cite{wu2019graph}. It combines graph convolution with dilated casual convolution to capture spatial-temporal dependecies. Moreover, it learns a latent graph structure using an attention-based method.
	\item STSGCN~\cite{song2020spatial}. It proposes a spatial-temporal synchronous graph convolutional network to capture localized spatial-temporal  correlations.
	\item AGCRN~\cite{bai2020adaptive}. It enhances GCNs using a node adaptive parameter learning module and a data adaptive graph generation module.
\end{itemize}

We reuse the results of DCRNN and STSGCN reported in the previous work~\cite{song2020spatial}. As for AGCRN and Graph WaveNet, we directly reuse the released codes without any modifications. The best parameters for all deep learning models are selected through a parameter-tuning process on the validation set.\\\\
\noindent\textbf{Evaluation Metrics} We use three widely-used evaluation metrics, i.e., Mean Absolute Error (MAE), Root Mean Squared Error (RMSE), and Mean Absolute Percentage Error (MAPE), to measure the performance of predictive models.

\begin{figure*}[htbp]
	\centering
\begin{tabular}{ccc}
	\includegraphics[width=.3\linewidth]{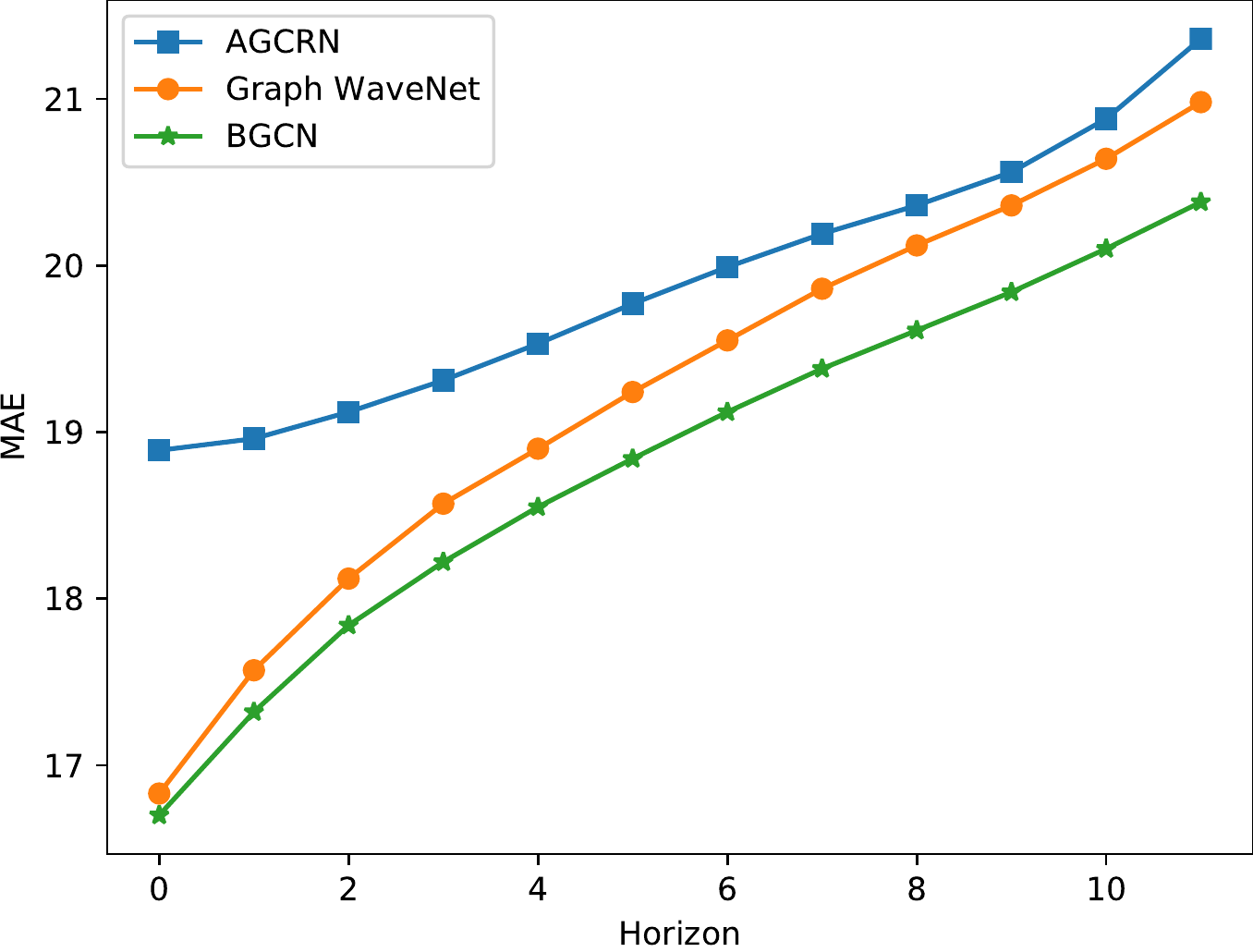} & 	\includegraphics[width=.3\linewidth]{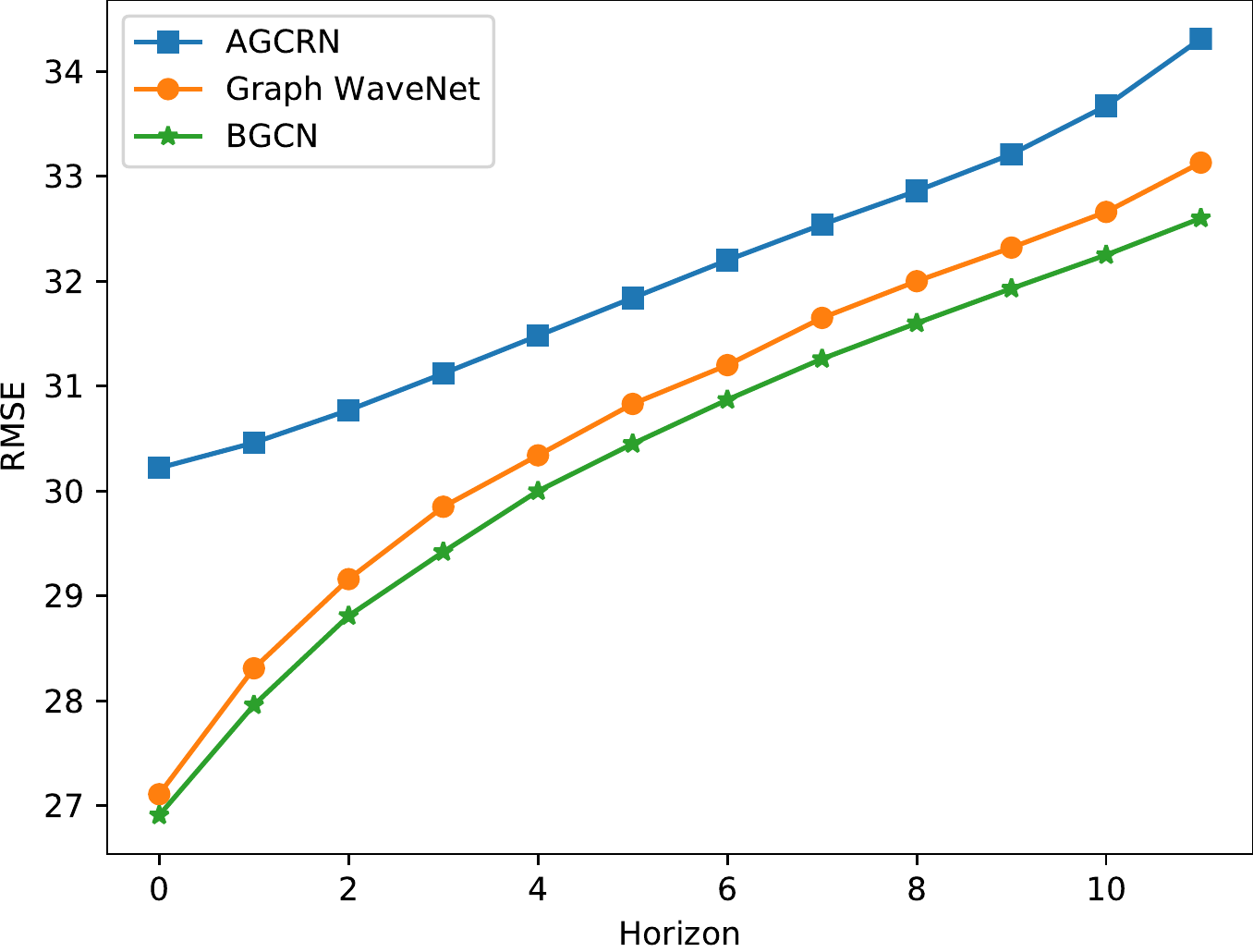}&
	\includegraphics[width=.3\linewidth]{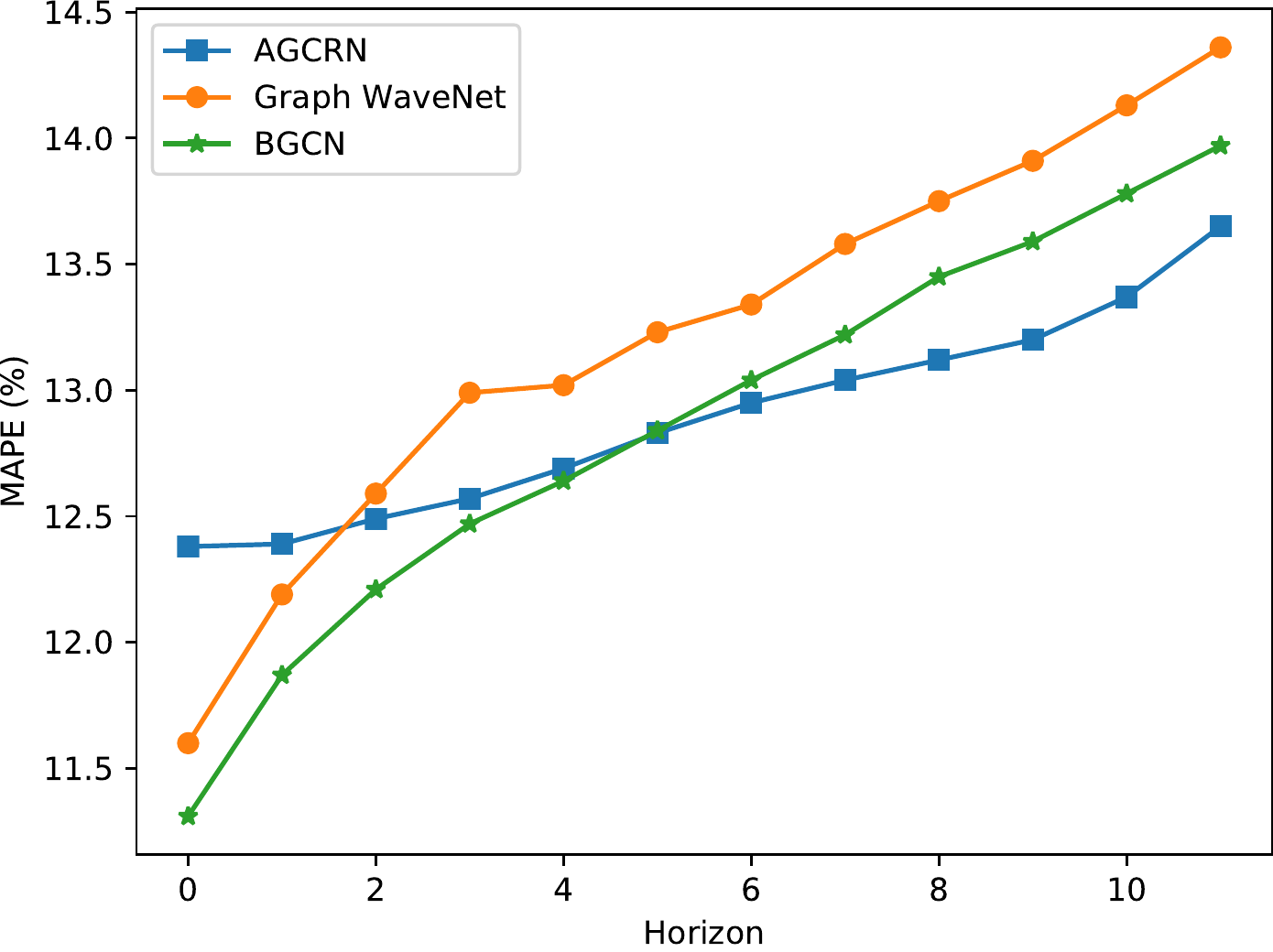}\\
	& (a) &\\
	\includegraphics[width=.3\linewidth]{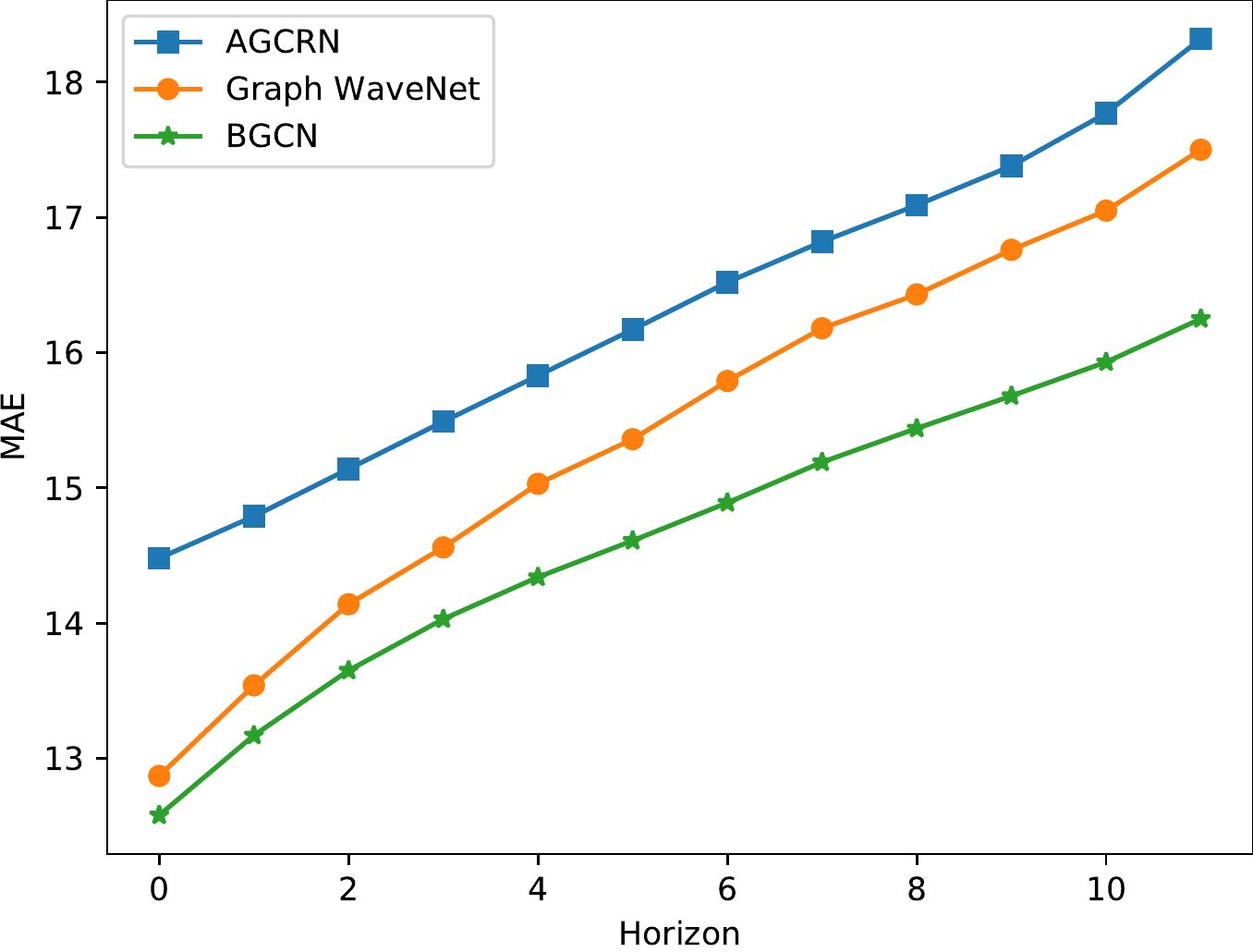} & 	\includegraphics[width=.3\linewidth]{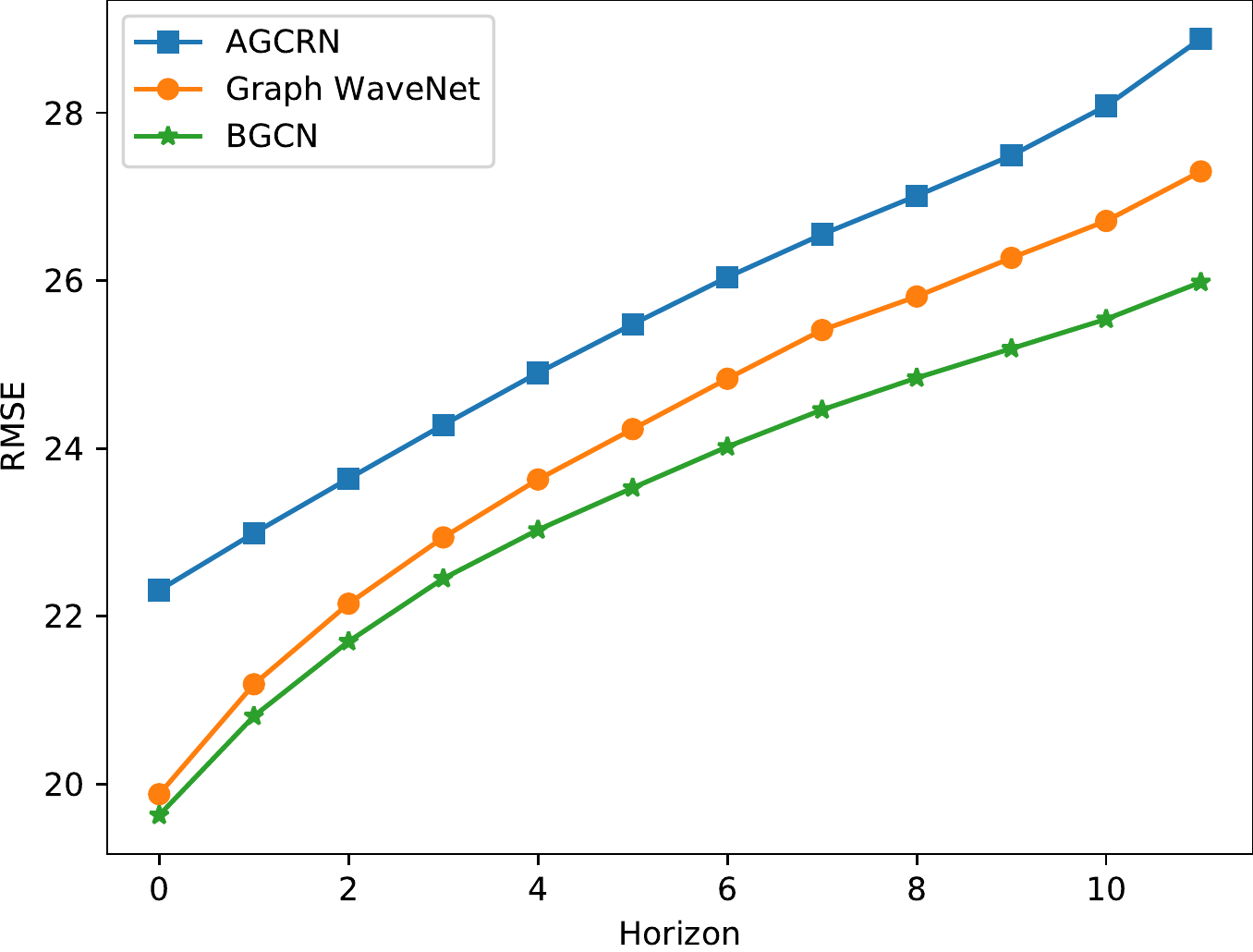}&
	\includegraphics[width=.3\linewidth]{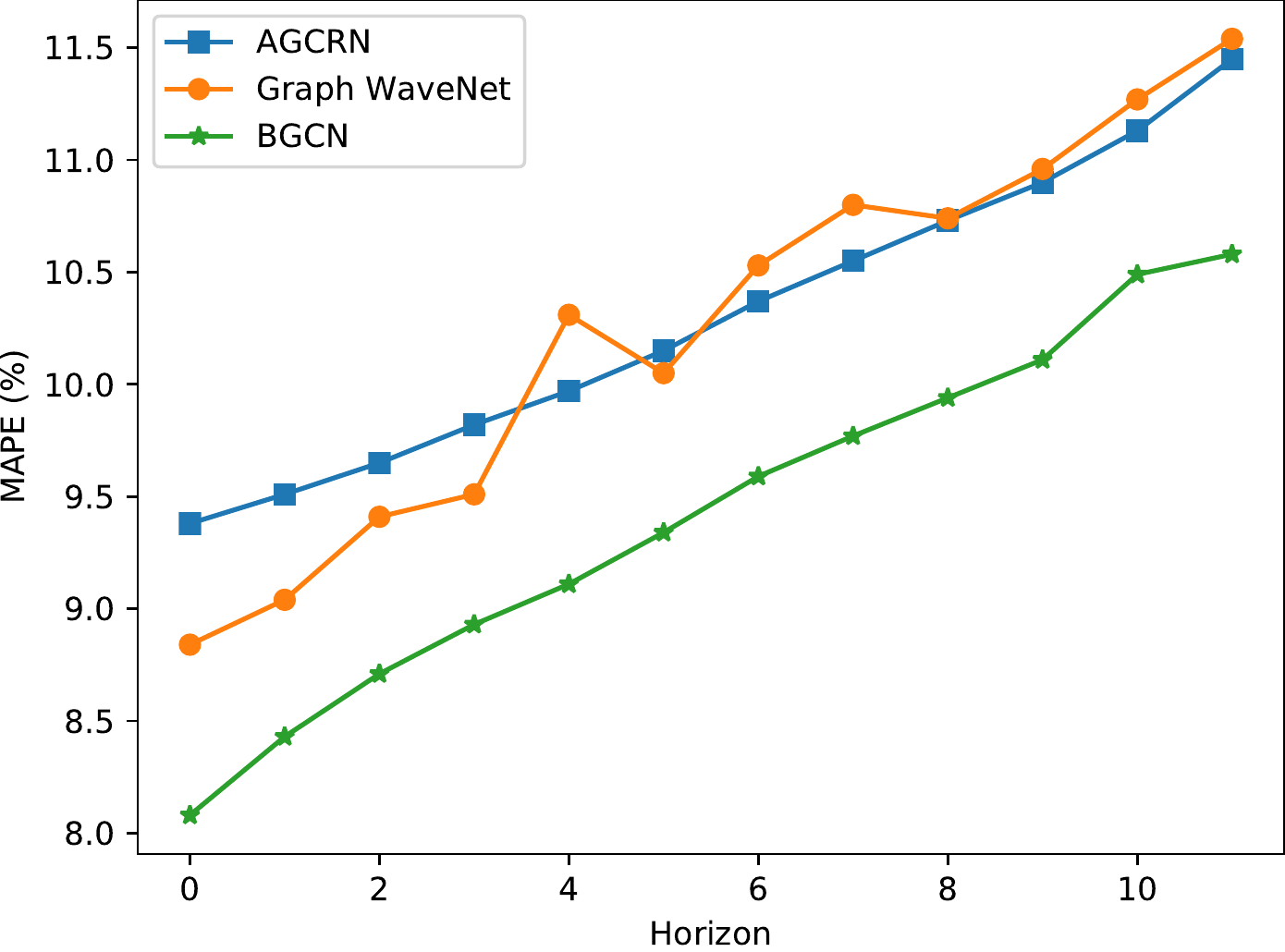}\\
	& (b) &\\
\end{tabular}
\caption{Prediction performance comparison at each horizon. (a) PeMS4 dataset (b) PeMS8 dataset.}
\label{fig:horizon}
\end{figure*}

\begin{figure*}[htbp]
	\centering
	\begin{tabular}{ccc}
		\includegraphics[width=.3\linewidth]{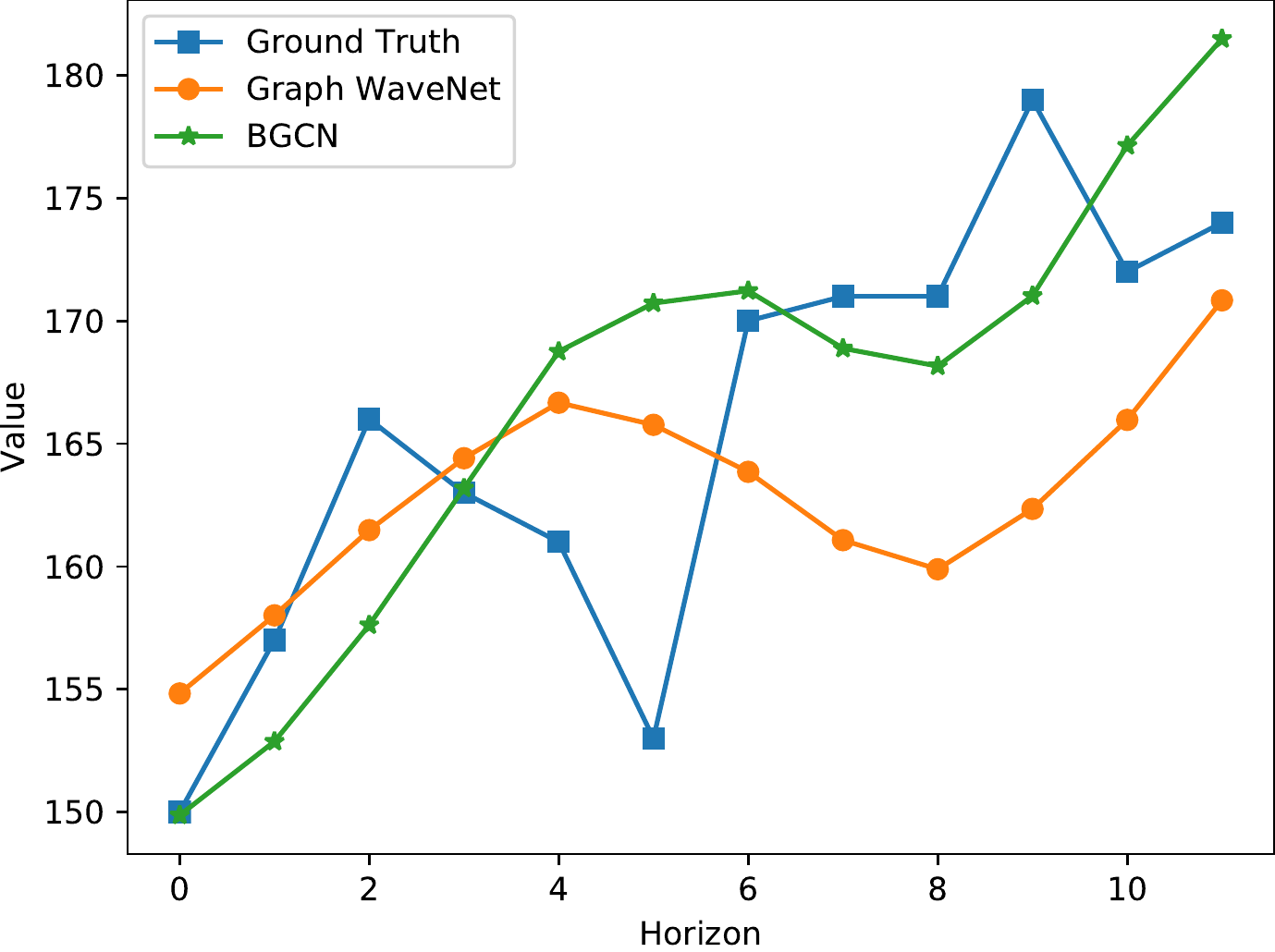} & \includegraphics[width=.3\linewidth]{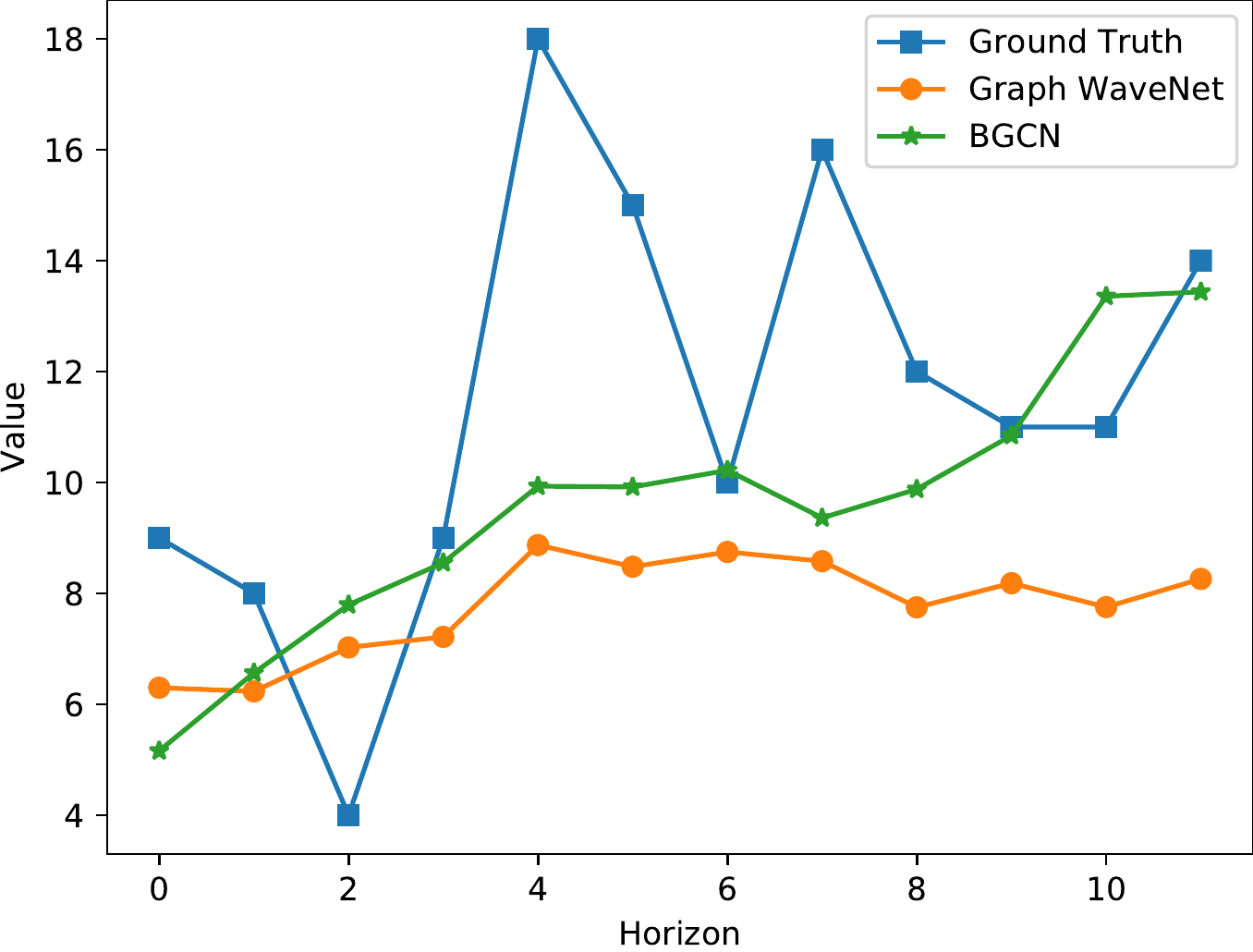} & \includegraphics[width=.3\linewidth]{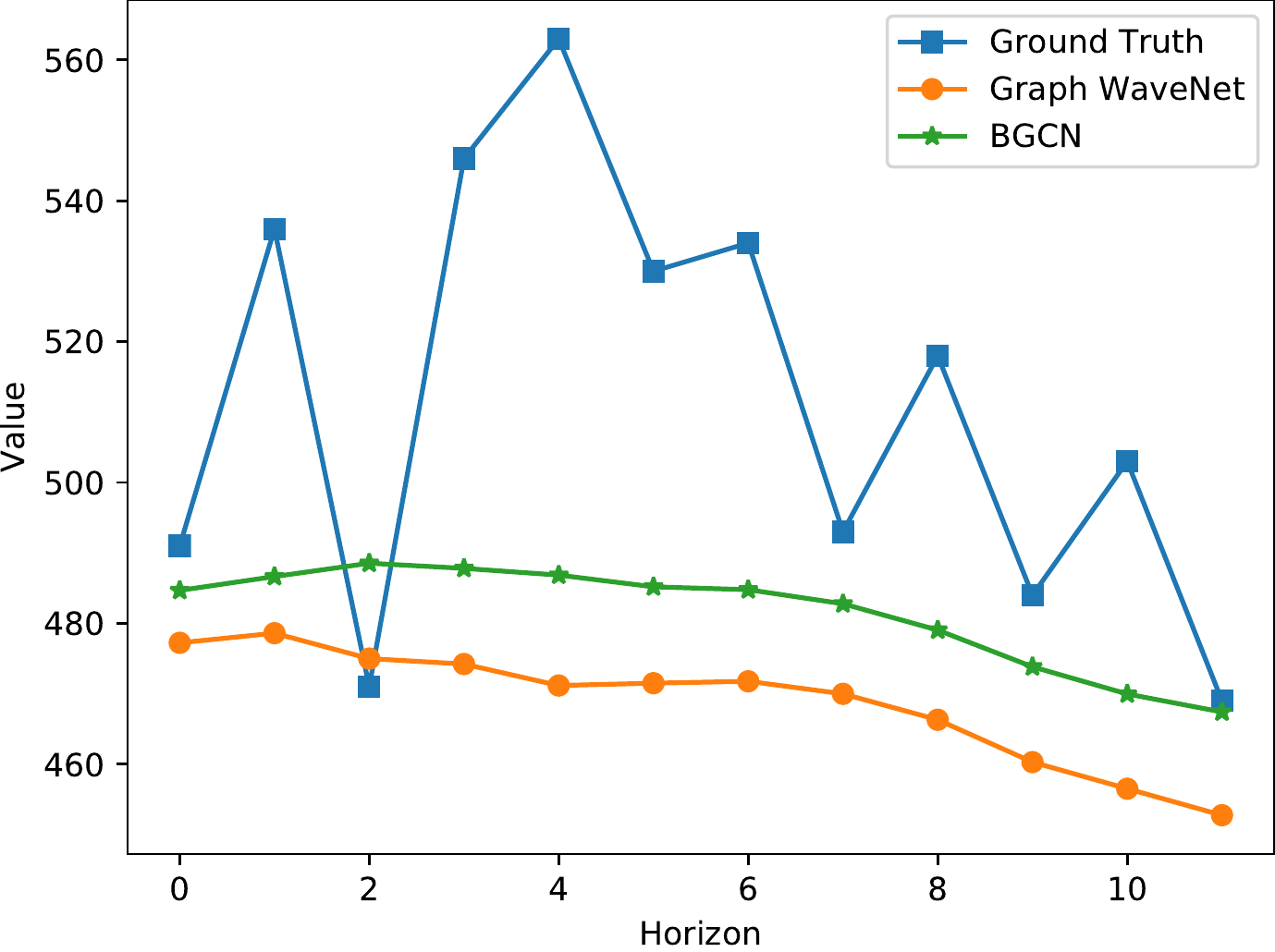}	\\
	\end{tabular}
	\caption{Visualization of some predicted results on the PeMS8 dataset.}
	\label{fig:trace}
\end{figure*}

\subsection{Overall Comparison}
Table ~\ref{tb:all results} presents the results, i.e., the averaged MAE, RMSE, and MAPE over 12 prediction horizons, of BGCN and eight representative baselines on five datasets. According to this table, we draw the following conclusions.
\begin{itemize}
	\item Traditional methods (i.e., HA and VAR) achieve competitive performance in the PeMS-Bay dataset with a small standard deviation, while perform poorly in the remaining datasets with large standard deviations. This reveals that  traditional methods have limited capability in handling the nonlinear and complex  traffic conditions due to the stationary assumption.	
	\item Compared to traditional approaches, learning-based methods achieve better prediction performance, especially the approaches that simultaneously take the spatial and temporal correlations into account.
	\item Compared to STGCN using heuristically defined graph structure, the methods that learn latent graph structures from traffic data, including Graph WaveNet, AGCRN, and BGCN, significantly improve the prediction performance. This indicates that road-network-topology based graph structure is not the optimal description of the spatial relationships between traffic conditions.
	\item Compared to Graph WaveNet and AGCRN, BGCN achieves the best performance on all datasets in terms of almost all evaluation metrics. This verify the excellent ability of  BGCN in modeling the graph structure.
\end{itemize}

Fig.\ref{fig:horizon} further shows the prediction performance at each horizon on the PeMSD4 and PeMS8 dataset. We have the following findings.
\begin{itemize}
	\item  The difficulty of traffic forecasting increases with the increase of prediction interval.
	\item Our method performs well in short-term and long-term traffic prediction and achieves the best performance at  all horizons in terms of almost all evaluation metrics (except for MAPE).
	\item Compared to  Graph WaveNet and AGCRN, BGCN shows a slower performance decline trend as the prediction interval increases.
\end{itemize}

\begin{figure*}[htbp]
	\begin{tabular}{ccc}
		\includegraphics[width=.3\linewidth]{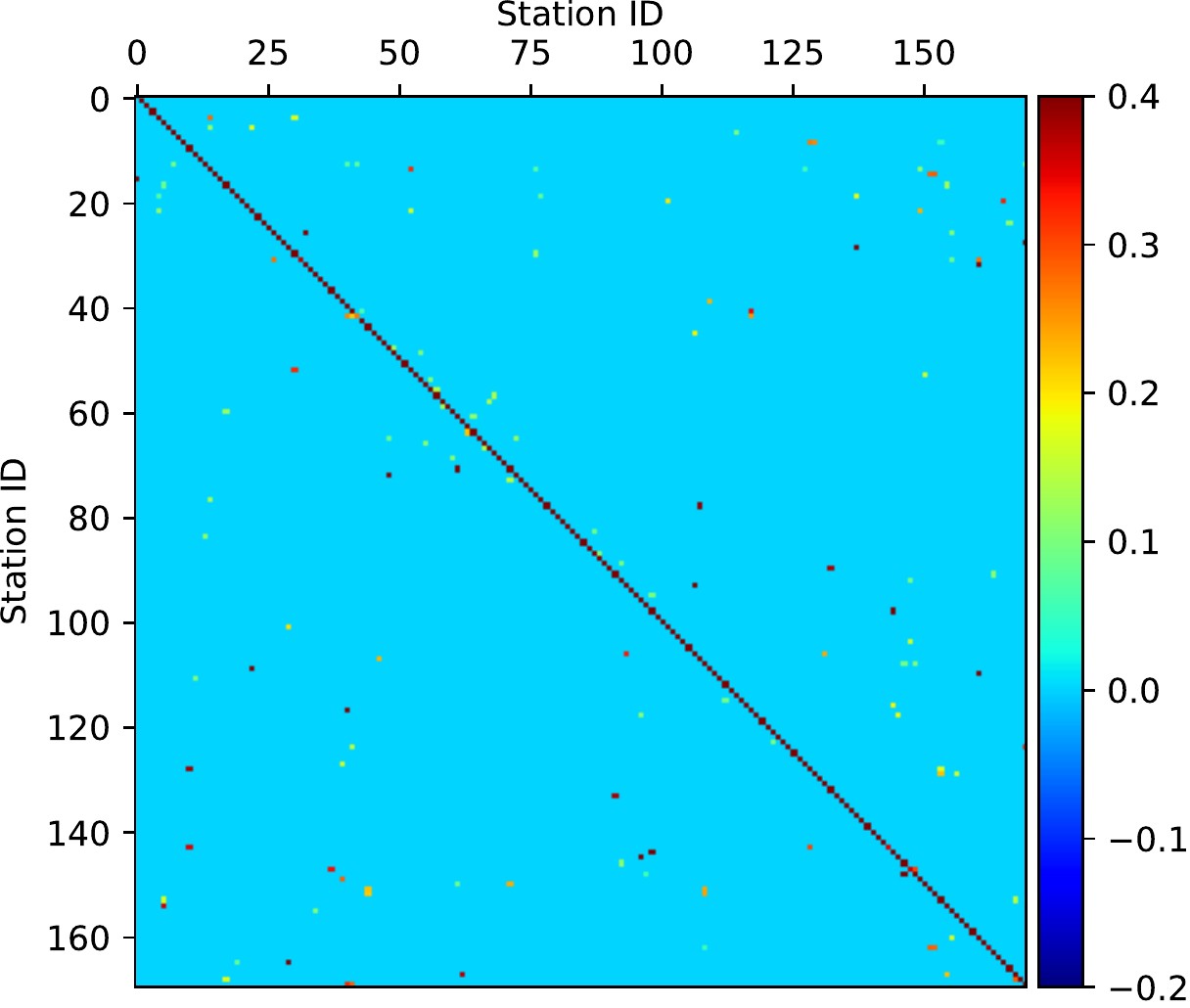} & \includegraphics[width=.3\linewidth]{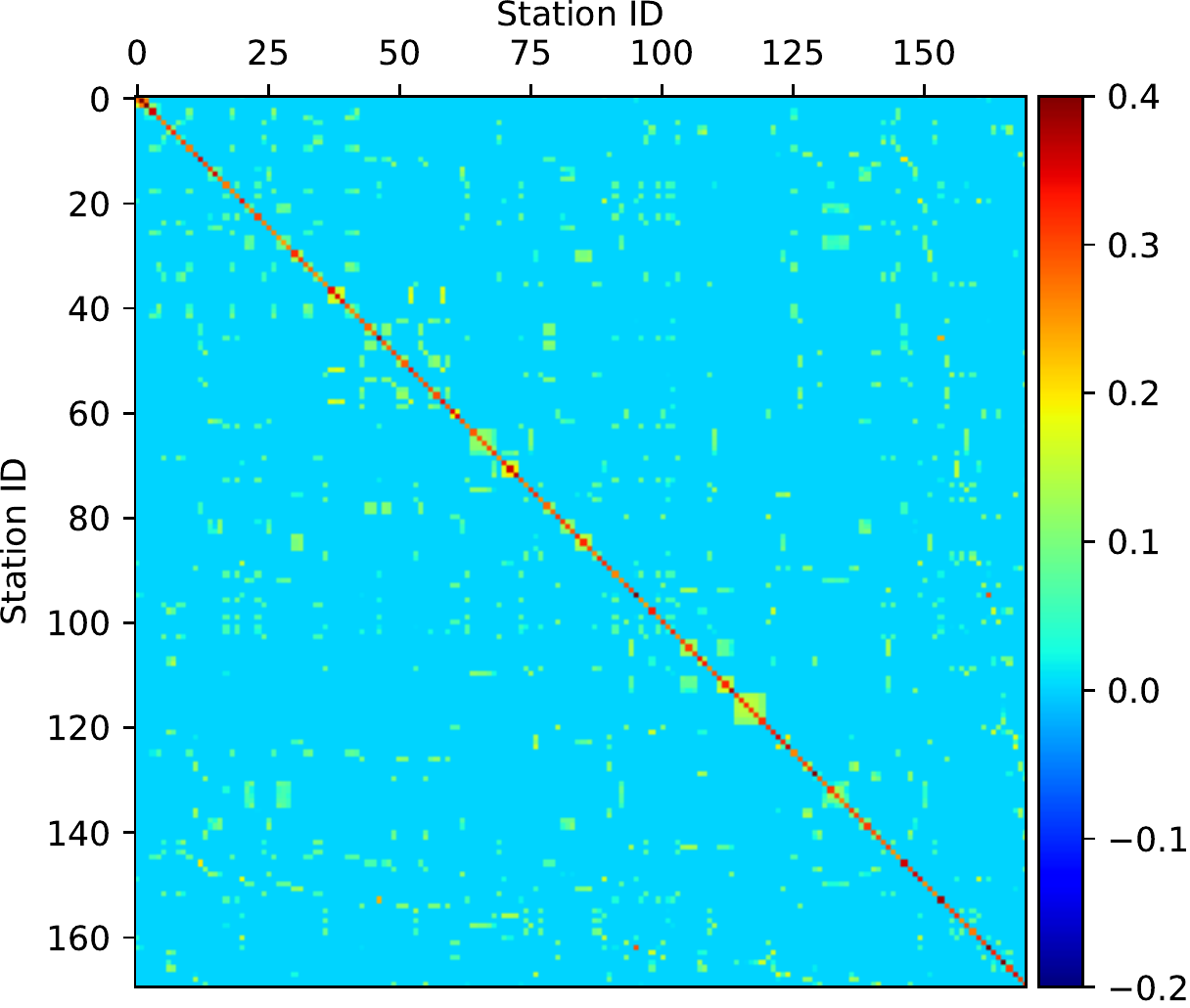} & \includegraphics[width=.3\linewidth]{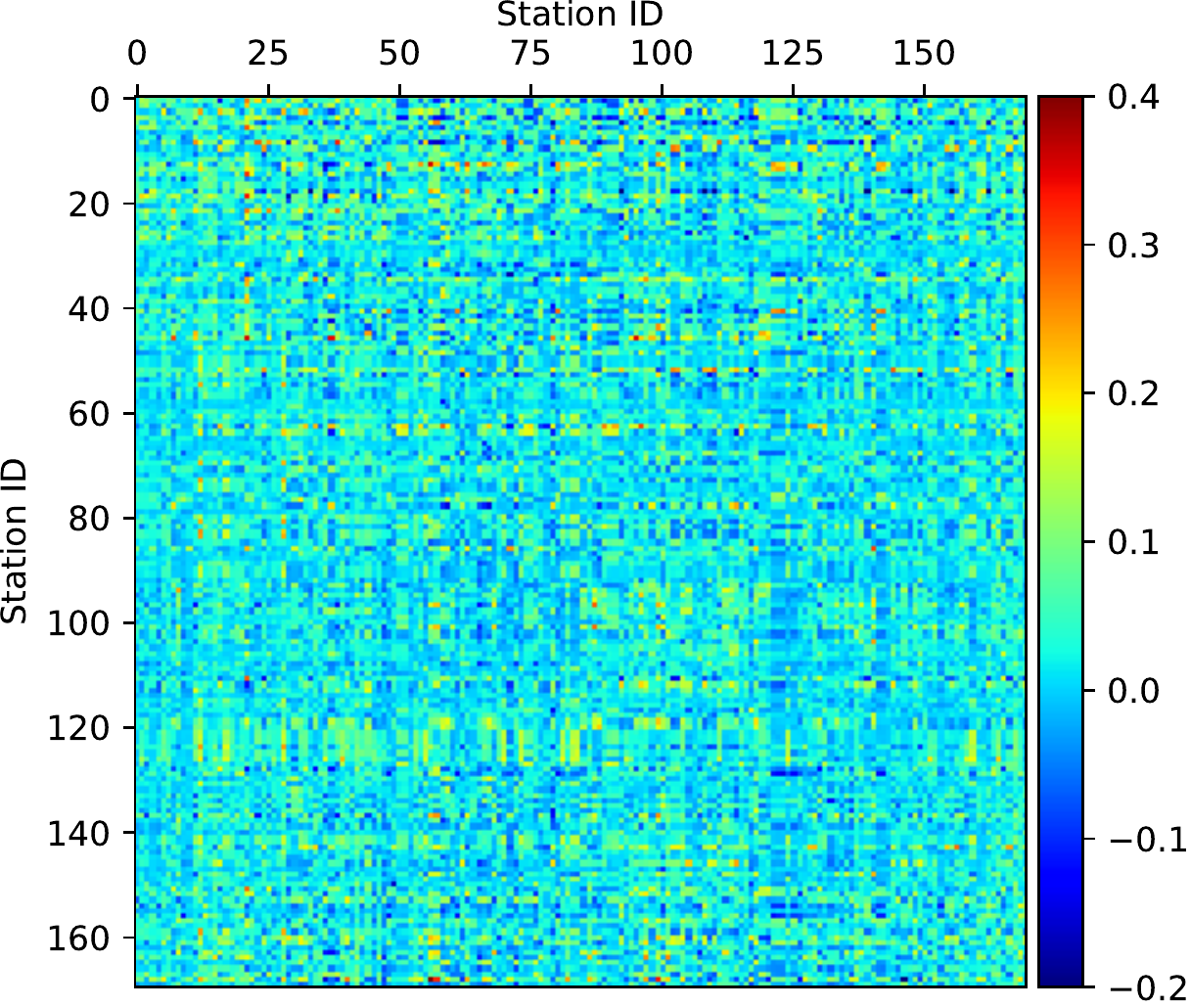}	\\
		(a) & (b) & (c)   \\
	\end{tabular}
	\caption{Illustration of adjacency matrices on the PeMS8 dataset. (a) The normalization observed adjacency matrix with self-loops $\tilde{\mathcal{A}}_{obs}$ (b) The constant adjacency matrix $\tilde{\mathcal{A}}_{\overline{g}}$ (c) The learned adjacency matrix $\tilde{\mathcal{A}}_{\overline{g}} + \phi$. }
	\label{fig:adj}
\end{figure*}

\begin{figure*}[htbp]
	\begin{tabular}{ccc}
		\includegraphics[width=.3\linewidth]{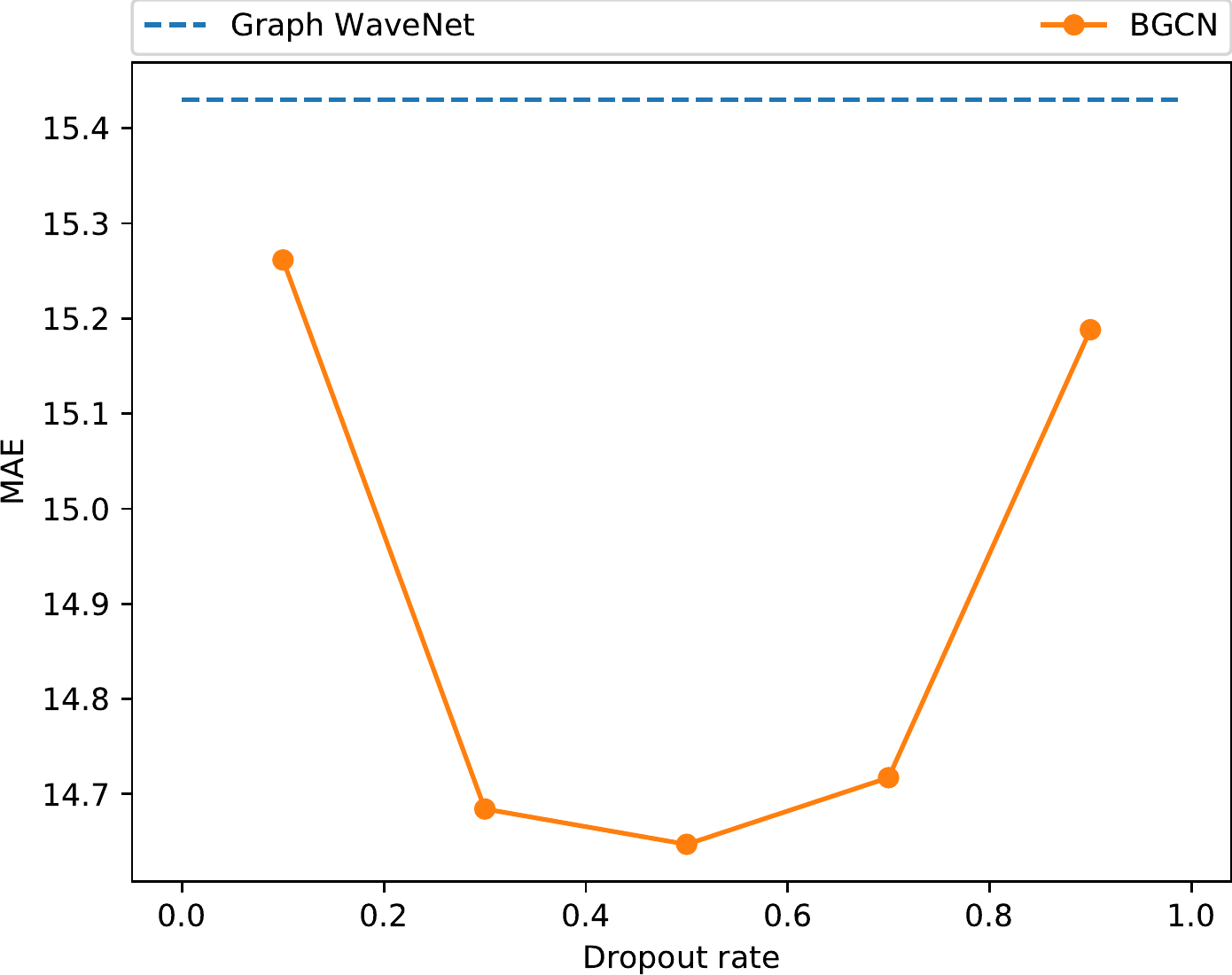} & \includegraphics[width=.3\linewidth]{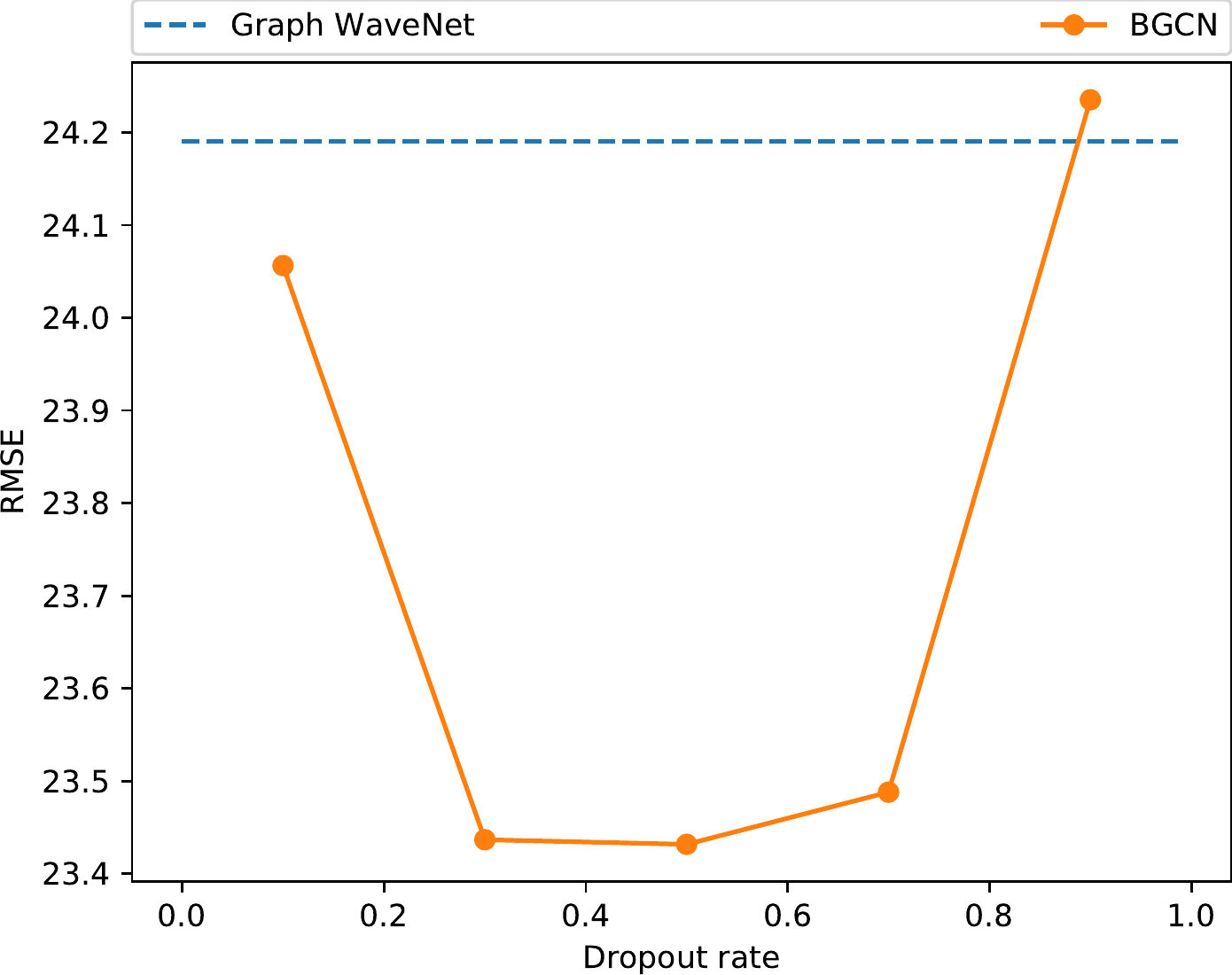} & \includegraphics[width=.3\linewidth]{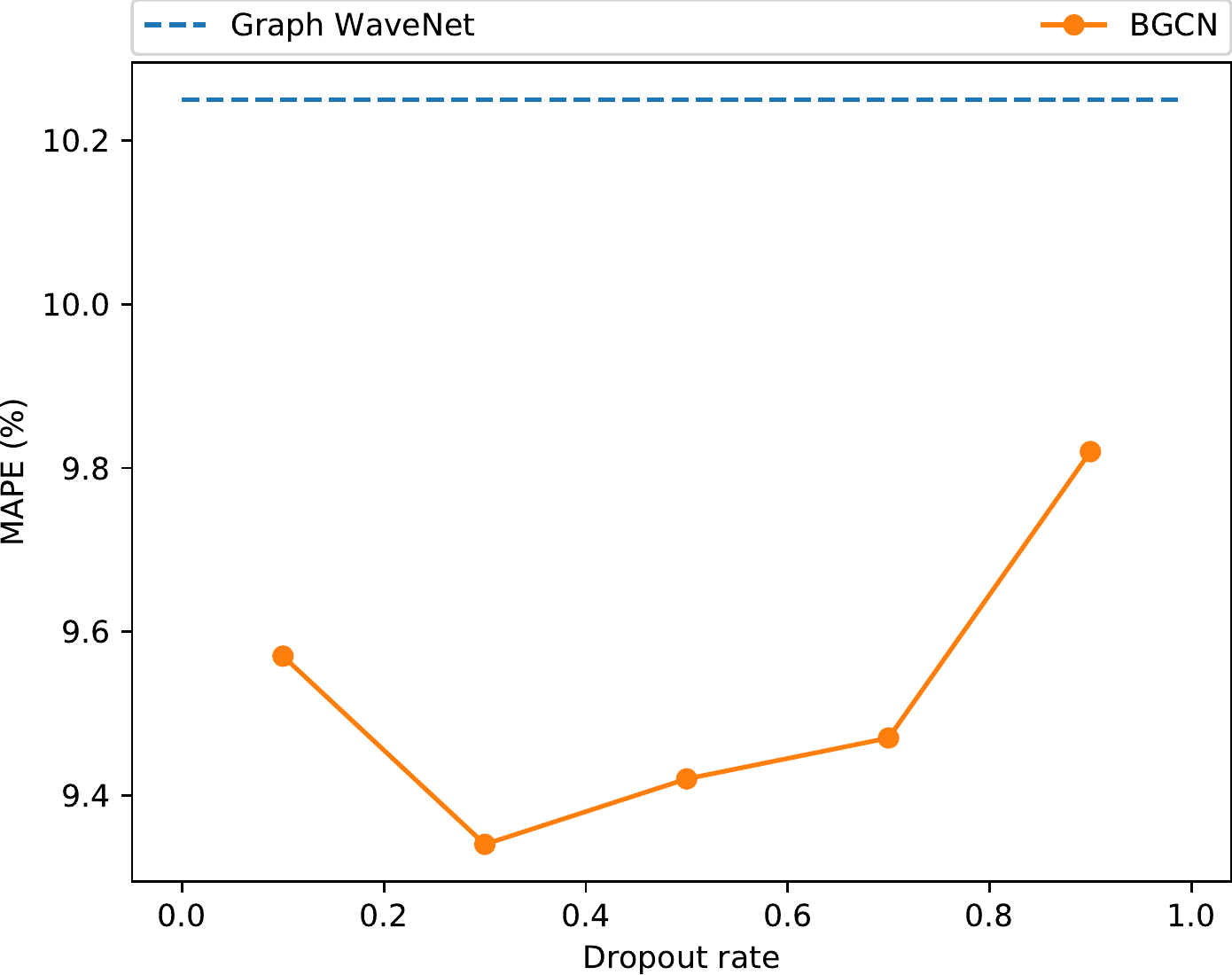}	\\
	\end{tabular}
	\caption{Investigation on the dropout rate in the Monte Carlo approximation.}
	\label{fig:parameter}
\end{figure*}

Fig.\ref{fig:trace} visualizes some predicted results of Graph WaveNet and BGCN on the PeMS8 dataset. We can observe that the estimated results of BGCN are more related to the ground truth values compared to those of Graph WaveNet. 

Table~\ref{tb:cost} reports the computation cost, i.e., the parametric size, training time per epoch, and inference time, of our method, STGCN, Graph WaveNet, and  AGCRN on the PeMSD8 dataset. We have the following observations.
\begin{itemize}
	\item AGCRN is inferior to STGCN and Graph WaveNet in training time per epoch and inference time, even with fewer parameters. This is mainly because RNNs sequentially process the input time series data unlike CNNs using parallel mechanism.
	\item Compared to STGCN, Graph WaveNet uses more parameters and has slower speed. Considering the significant performance improvement (as shown in Table~\ref{tb:all results}), this is moderate.

	\item Compared to Graph WaveNet, BGCN uses fewer parameters while achieves superior prediction performance. Besides, BGCN slightly shortens the training and inference time.
\end{itemize}

\begin{table}[h]
	\caption{the computation cost on the pems8 dataset. ``dim" means the dimension of node embedding. ``*'' denotes the setting of node embedding  for the pems4 dataset.}
	\centering
	\begin{tabular}{@{}cccc@{}}
		\toprule
		Model          & \# Parameters & Training Time & Inference Time\\ \midrule
		STGCN          & 211596        & 12.35 s        &  3.6 s    \\ \midrule
		Graph WaveNet & 305228  &  31.20 s& 5.4 s\\ \midrule
		AGCRN (dim = 2)  & 150386        & 33.88 s     &   13.75 s       \\ \midrule
		AGCRN$^*$ (dim = 10) & 748810        & 35.56 s        &14.28 s     \\ \midrule
		BGCN & 256076        & 33.14 s       &  4.4 s      \\
		\bottomrule
	\end{tabular}
	\label{tb:cost}
\end{table}

\begin{table}[htbp]
	\caption{ablation study on the pems8 dataset.}
	\centering
	\begin{tabular}{@{}lccc@{}}
		\toprule
		Model Setting         & MAE & RMSE & MAPE (\%)\\ \midrule
		BGCN          & 14.65       &    23.43    &  9.42 \\ \midrule
		-----\ Uncertainty & 16.45 &  25.49 & 10.95 \\
		-----\ Learnable Adjacency Matrix & 17.49       & 27.60     &   11.15   \\ 
		-----\ Constant Adjacency Matrix & 15.06       & 23.96        & 9.53     \\ 
		\bottomrule
	\end{tabular}
	\label{tb:ablation study}
\end{table}
\subsection{Ablation study}
Table~\ref{tb:ablation study} presents the results of the ablation study on the PeMS8 dataset. According to this table, we have the following findings.
\begin{itemize}
	\item The prediction error of our method increases a lot when the learnable adjacency matrix is removed. This indicates  learning the latent graph structure from traffic data is necessary for traffic forecasting.
	\item Introducing the uncertainty into the graph structure significantly improves the performance of traffic prediction. 
	\item The prediction performance drops when we learn the underlying graph structure from scratch (i.e., removing the constant adjacency matrix). This means  the prior of the observed topology of the road network is helpful for finding a more accurate description of spatial correlations.
\end{itemize}

\begin{figure}[htbp]
	\begin{tabular}{c @{\hspace{.5em}} c}
		\includegraphics[width=.5\linewidth]{material/pems08_full_adj.pdf} & 	\includegraphics[width=.46\linewidth]{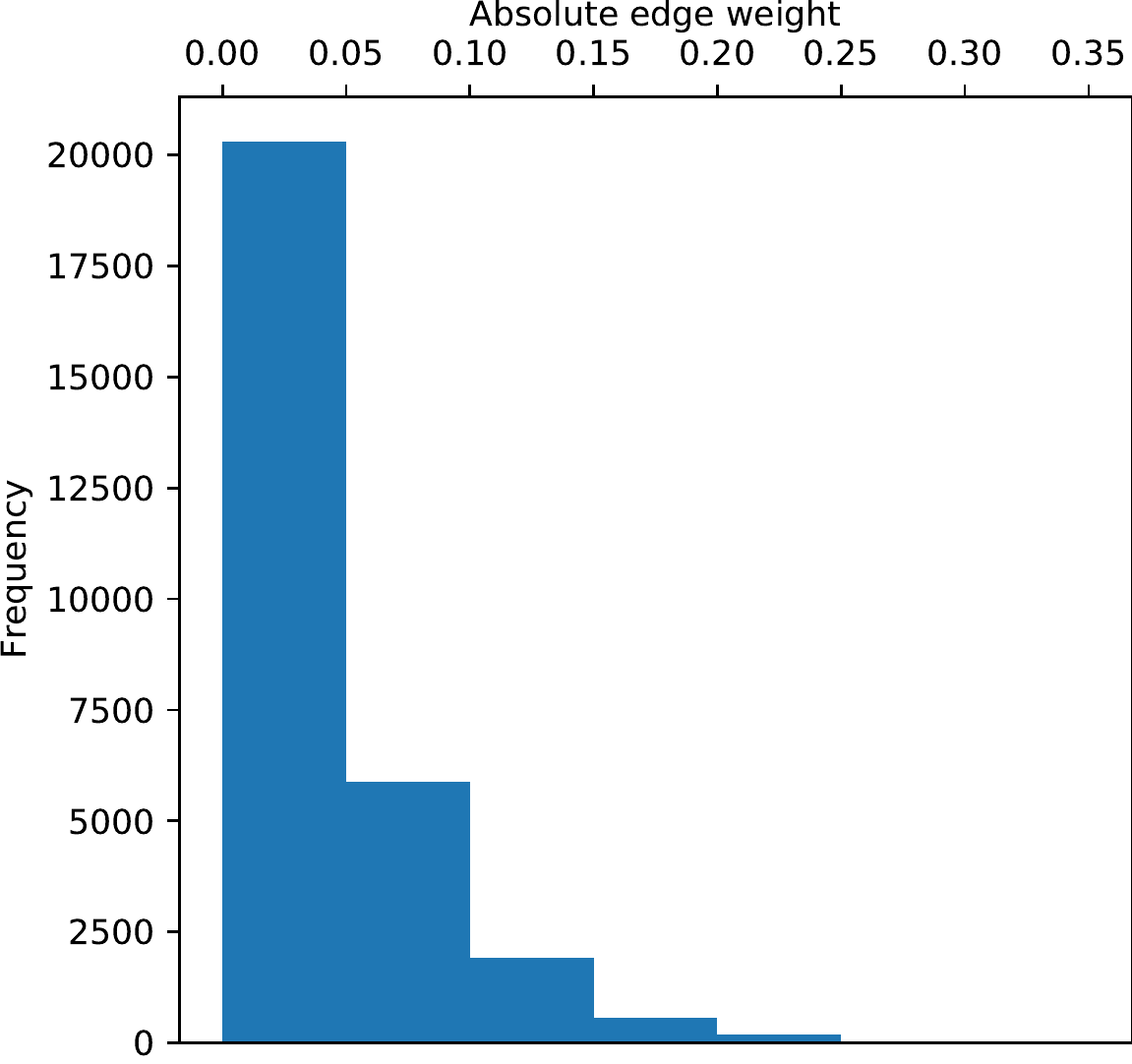}\\
		\multicolumn{2}{c}{(a) }\\
		\includegraphics[width=.5\linewidth]{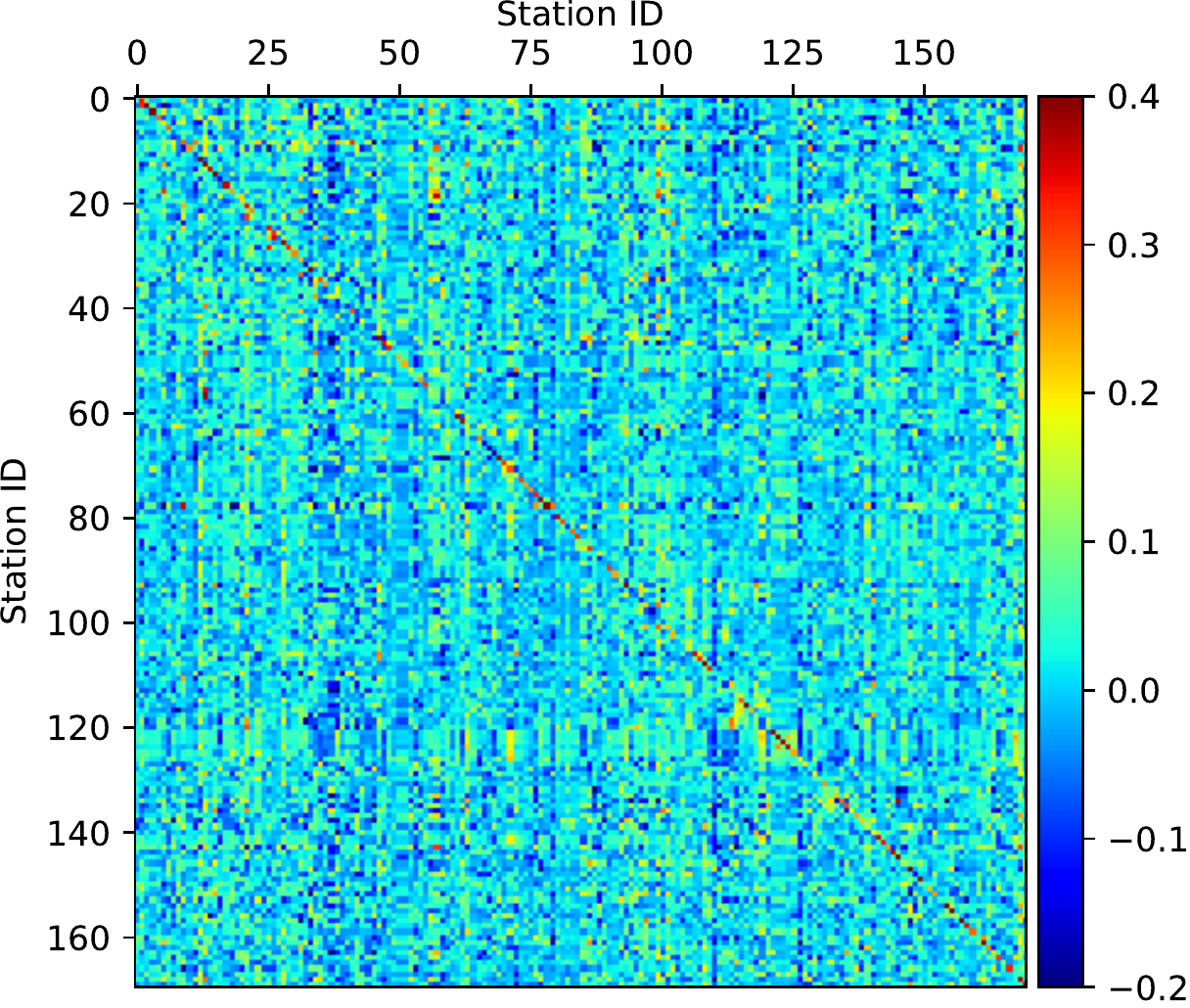} & 	\includegraphics[width=.46\linewidth]{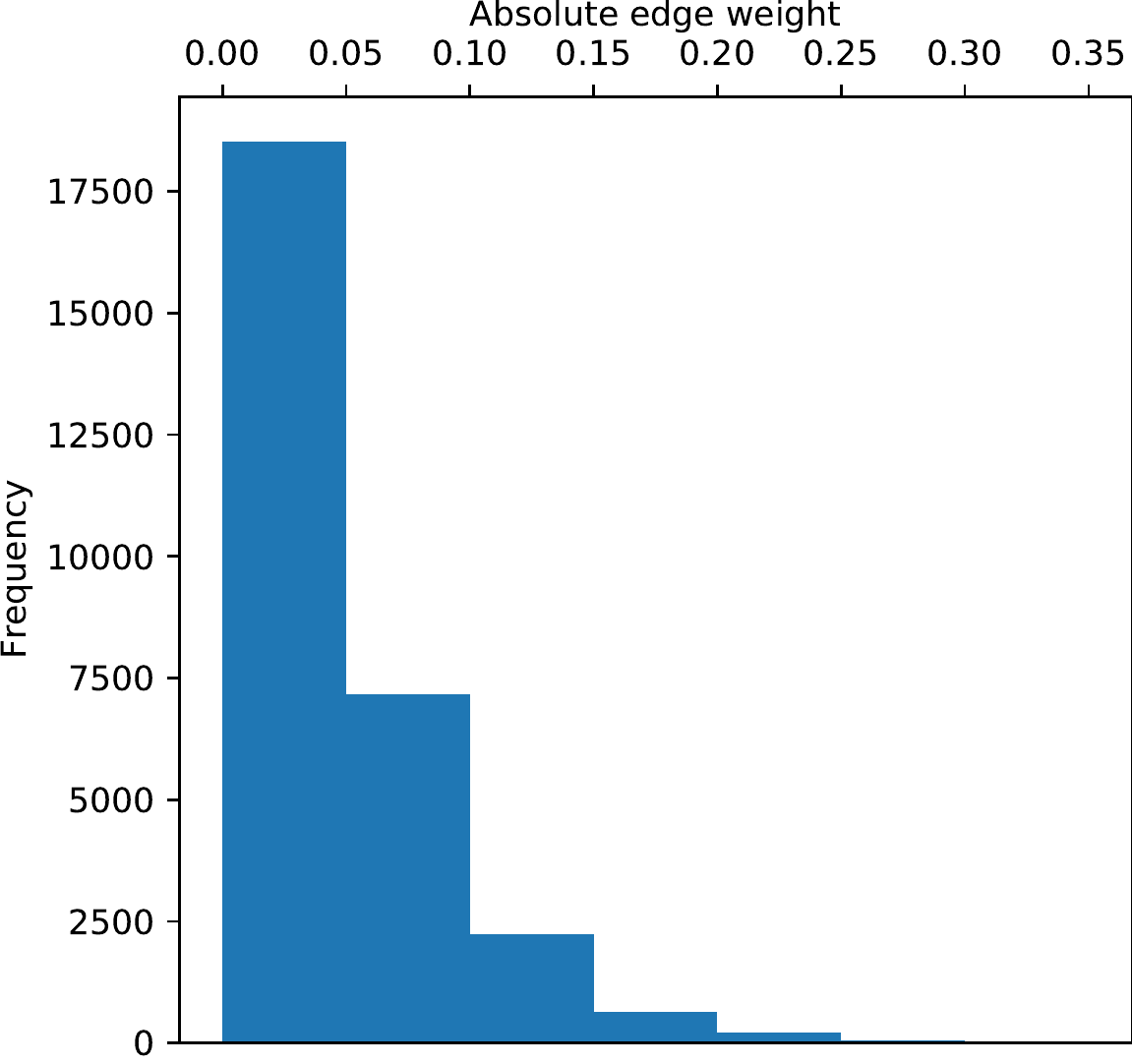} \\
		\multicolumn{2}{c}{(b)}\\
	\end{tabular}
	\caption{Impact of uncertainty on graph structure learning on the PeMS8 dataset. (a) Learning graph structure with uncertainty (b) Learning graph structure without uncertainty.  }
	\label{fig:uncertainty}
\end{figure}
To further figure out the benefit of introducing uncertainty into the graph structure, we visualize the learned adjacency matrix and the histogram of absolute edge weights in Fig.~\ref{fig:uncertainty}. We can observe that the adjacency matrix learned without uncertainty has larger scale than that of BGCN. This means that introducing uncertainty into the graph structure acts as a regularizer and tends to sparse the adjacency matrix.

We also show the observed adjacency matrix, the constant adjacency matrix, and the learned adjacency matrix in Fig.~\ref{fig:adj}. We have the following observations. 
\begin{itemize}
	\item The observed adjacency matrix $\tilde{\mathcal{A}}_{obs}$  is typically sparse, which may miss some important spatial correlations between traffic conditions.
	\item Compared to $\tilde{\mathcal{A}}_{obs}$ , the constant adjacency matrix $\tilde{\mathcal{A}}_{\overline{g}}$ contains more edges and is symmetric.
	\item Among presented adjacency matrices, the learned adjacency matrix $\tilde{\mathcal{A}}_{\overline{g}} + \phi$ has most edges and is asymmetric. Furthermore, the negative edge weight is also included.
\end{itemize}
%
%
\begin{table}
	\centering
	\caption{exploration of generalization of bgcn on the pems8 dataset.}
	\begin{tabular}{l|cc|cc}
		\toprule
		Metric & STGCN & STGCN-BGCN & AGCRN & AGCRN-BGCN \\

		\midrule
	 MAE & 16.98& \textbf{16.70 }& 16.29 & \textbf{15.63}\\
		RMSE & 26.58& \textbf{25.85} & 25.66 & \textbf{24.79}\\
	MAPE (\%) & 11.58 & \textbf{11.03} & \textbf{10.32} & 10.38\\
		\bottomrule
	\end{tabular}
	\label{tb:generalization}
\end{table}
\subsection{ Parameter Experiments}
One key parameter in BGCN is the dropout rate in the Monte Carlo approximation. Fig.~\ref{fig:parameter} shows the effects of different dropout rates to BGCN on the PeMSD8 dataset.
We can observe that BGCN improves the prediction performance of Graph WaveNet at almost all the tested dropout rates, which shows the robustness of BGCN. Moreover, BGCN achieves the best performance when the dropout rate is assigned to 0.5. In addition, both an excessively small and large dropout rate will weaken the prediction performance.

\subsection{Generalization Experiments}
Our proposed BGCN is a plug-and-play module. We generalize it to some other graph-based traffic prediction networks, i.e., STGCN and AGCRN. The results are listed in Table~\ref{tb:generalization}. We can observe that BGCN-applied STGCN and AGCRN exceed the original version by a noticeable margin in terms of MAE and RMSE. In addition, we can find that BGCN slightly impairs the MAPE performance of AGCRN. One possible reason is that the used objective function directs the traffic prediction network to develop towards minimizing MAE instead of minimizing MAPE.



\section{Conclusion and Future Work}
In this paper, we propose a Bayesian Graph Convolutional Network for traffic prediction. It introduces the information of traffic data and uncertainty into the graph structure using a Bayesian approach. Moreover, it is a plug-and-play module for graph-based traffic prediction networks. Experimental results on five real-world datasets verify the effectiveness and the generalization ability of BGCN in traffic prediction. In the future, we focus on  extending BGCN to other spatio-temporal time series forecasting tasks, such as forecasting ride demand.


%

\section*{Acknowledgment}
This work was supported in part by NSFC under Grant U1908209, 61632001 and the National Key Research and Development Program of China 2018AAA0101400.




%
\bibliographystyle{IEEEtran}
\bibliography{ref}

%
%

%

%
%
%




\end{document}